\documentclass[acmsmall]{acmart}

\newcommand{\mypara}[1]{\vspace*{0.75ex}\noindent{\bf #1}~}

\newcommand{\be}{\begin{itemize}}  
\newcommand{\ee}{\end{itemize}}  
\newcommand{\bn}{\begin{enumerate}}  
\newcommand{\en}{\end{enumerate}}  
\newcommand{\bc}{\begin{center}}  
\newcommand{\ec}{\end{center}}  
\newcommand{\bl}{\begin{flushleft}}  
\newcommand{\el}{\end{flushleft}}  
\newcommand{\bq}{\begin{quote}}  
\newcommand{\eq}{\end{quote}}

\newcommand{\bmp}{\begin{minipage}}  
\newcommand{\emp}{\end{minipage}}

\newtheorem{instance}{Example}
\newcommand{\bexa}{\begin{instance}\rm}
\newcommand{\eexa}{\end{instance}}
\newcommand{\eeeexa}{\end{itemize}\end{instance}}
\newcommand{\eeeeexa}{\rule[-0.1mm]{1.0mm}{3mm}\end{itemize}\end{itemize}\end{example}}

\usepackage{tikz}
\tikzset{every picture/.style={/utils/exec={\sffamily}}}
\usepackage{graphicx,subcaption}
\usetikzlibrary{positioning}
\usepackage{algorithmic}
\usepackage{textcomp}
\usepackage{xcolor}
\usepackage{color,soul}
\usepackage{mathtools}
\usepackage{caption}
\DeclarePairedDelimiterX{\infdivx}[2]{[}{]}{
  #1\;\delimsize\|\;#2%
}
\newcommand{\KL}{\text{KL}\infdivx}
\newcommand{\pure}{\textsc{PURE}}

\AtBeginDocument{%
  \providecommand\BibTeX{{%
    \normalfont B\kern-0.5em{\scshape i\kern-0.25em b}\kern-0.8em\TeX}}}

\setcopyright{acmcopyright}
\acmJournal{TOIT}
\acmYear{2022} \acmVolume{1} \acmNumber{1} \acmArticle{1} \acmMonth{1} \acmPrice{15.00}\acmDOI{10.1145/3471187}

\acmBooktitle{ACM Transactions on Internet Technology}
\acmPrice{15.00}
\acmISBN{978-1-4503-XXXX-X/18/06}

\begin{document}

\title{Uncertainty-aware Personal Assistant for Making Personalized Privacy Decisions}

\author{G\"on\"ul Ayc{\i}}
\affiliation{\institution{Bogazici University}
    \country{Turkey}}
\email{gonul.ayci@boun.edu.tr}

\author{Murat Sensoy}
\affiliation{\institution{Ozyegin University}
    \country{Turkey}}
\email{drmuratsensoy@gmail.com}

\author{Arzucan \"Ozg\"ur}
\affiliation{\institution{Bogazici University}
    \country{Turkey}}
\email{arzucan.ozgur@boun.edu.tr}

\author{P{\i}nar Yolum}
\affiliation{\institution{Utrecht University}
    \country{The Netherlands}}
\email{p.yolum@uu.nl}

\begin{abstract}
Many software systems, such as online social networks enable users to share information about themselves. While the action of sharing is simple, it requires an elaborate thought process on privacy: what to share, with whom to share, and for what purposes. Thinking about these for each piece of content to be shared is tedious. Recent approaches to tackle this problem build personal assistants that can help users by learning what is private over time and recommending privacy labels such as private or public to individual content that a user considers sharing. However, privacy is inherently {\it ambiguous} and highly {\it personal}. Existing approaches to recommend privacy decisions do not address these aspects of privacy sufficiently. Ideally, a personal assistant should be able to adjust its recommendation based on a given user, considering that user's privacy understanding. Moreover, the personal assistant should be able to assess when its recommendation would be uncertain and let the user make the decision on her own. Accordingly, this paper proposes a personal assistant that uses evidential deep learning to classify content based on its privacy label. An important characteristic of the personal assistant is that it can model its uncertainty in its decisions explicitly, determine that it does not know the answer, and delegate from making a recommendation when its uncertainty is high. By factoring in user's own understanding of privacy, such as risk factors or own labels, the personal assistant can personalize its recommendations per user. We evaluate our proposed personal assistant using a well-known data set. Our results show that our personal assistant can accurately identify uncertain cases, personalize them to its user's needs, and thus helps users preserve their privacy well.

\end{abstract}


\maketitle

\section{Introduction}
\label{intro}
Collaborative systems, such as online social networks (OSNs), enable users to share content with others. With the plethora of online content that is being shared, users are faced with the task of {\it managing} their privacy. Whenever a user is sharing content, she needs to think through whom the content is shared with, whether the content contains elements that would jeopardize her privacy, and so on. Some systems provide settings to configure sharing behavior, e.g., images can be shared only with friends. However, not all images are the same. For example, a user might be comfortable sharing a landscape image publicly, while she might prefer a family image to be shown only to friends. With current systems, identifying whether an image contains certain aspects that could be considered private is left to the user.
Moreover, the content may be shared by the user herself and others. For a user to decide whether her privacy is being violated, she needs to check the contents related to her individually. This is obviously time-consuming and error-prone. Ideally, a personal assistant (software) could help the user make decisions by signaling whether the content could be private.

Personal assistants help their users make decisions to ease their online interactions. Personal assistants have been used to help users in various tasks, including time management~\cite{myers2007intelligent}, smart homes~\cite{smart-personal-assistant}, voice-assistance~\cite{hauswald2015sirius}, and so on. 
Recently, personal assistants have been used for helping users manage their privacy online. K\"okciyan and Yolum~\cite{priguard-tkde} develop personal assistants to detect privacy violations in OSNs on behalf of their users. They assume that the personal assistant has access to users' privacy preferences through elicitation. Using these preferences and a domain ontology, the personal assistant computes whether others in the OSN share content about the user against her preferences. Kekulluoglu \textit{et al.}~\cite{kekulluoglu-18} and Such and Rovatsos~\cite{such-rovatsos-16} develop techniques to help users reach privacy decisions when a content being shared is owned by multiple users, such as a group image. Both approaches assume that the personal assistants of the users know the privacy preferences of the users and then they apply negotiation techniques to enable the personal assistants to reach a sharing decision that both users are comfortable with.

Because many approaches depend on using users' privacy preferences, there is a tremendous need to learn users' privacy preferences accurately. If a personal privacy assistant can represent the privacy expectations well, then these privacy assistants can help the users in privacy dealings, such as warning the user when the user attempts to share a private content, negotiate with other users on behalf of the user, and so on.   

While learning privacy preferences of a user resembles a classical machine learning problem, there are two properties of privacy that make the problem difficult. First, privacy by definition is ambiguous, making it challenging to specify. This makes the pattern that is searched malleable. Second, the users themselves are not always certain about their own privacy preferences and may change their preferences based on other motives~\cite{acquisti2005privacy}. For these reasons, using a traditional predictive model is unreliable as the cost of making a wrong privacy decision is high.

Ideally, the personal privacy assistant should adhere to the followings properties:
\begin{itemize}
    \item Unobtrusive: The privacy assistant should learn from the sharing behavior of the user without interrupting the user (e.g., asking the user what to share or not) as well as without requiring additional information about the user or the content, such as age or occupation of the user or the tags of a content. Thus, the privacy assistant should only consult the user if necessary.
    \item Uncertainty-aware: As mentioned above, privacy decisions are many times ambiguous. A personal privacy assistant may not always be able to decide if a content is private or not for the user. The assistant should be aware of this uncertainty and be able to say “I don't know” rather than making an uncertain decision. Hence, it should let the user know that it is uncertain and delegate the decision back to the user.
    \item Personalized: There are two aspects of personalization that are important for the personal assistant to consider. First, to be able to understand the privacy expectations of its own user. This is important because privacy is subjective and what one user considers private might not be private for another user. Second, each user has a different {\it risk} associated with making a wrong decision. The risk here refers to classifying a content as private when it should have been public and vice versa. For example, a user might prefer that the personal assistant be risk-averse and classify a content as public when there is even a slight chance that the user would prefer it private.  
\end{itemize}

Existing privacy personal assistants that learn users' privacy preferences do not address the uncertainty of their predictions while making decisions~\cite{tonge2020image,kurtan2021assisting,misra2017pacman} (see Section~\ref{sec:related} for details). The idea of considering risk to personalize decisions has been used before but not been coupled with privacy decisions as we have done here~\cite{sensoy2021misclassification}. Accordingly, this paper proposes a personal privacy assistant (\pure) that helps its user to make privacy decisions in a personalized way, taking into account the ambiguity of privacy predictions. An important aspect of \pure~is that it explicitly calculates the uncertainty of its decisions using evidential deep learning (EDL)~\cite{sensoy2018evidential}, which quantifies the predictive uncertainty of deep neural networks. When \pure~is uncertain of its decisions, it delegates the prediction back to the user when its uncertainty is high. \pure~uses publicly annotated data set to create an initial model for privacy but also factors in {\it persona} of a user: person's understanding of risk, personally labeled data, and when she should be consulted. In this way, \pure~behaves differently for each user to minimize the user's perceived risk of privacy violations. Moreover, \pure~does not need to have access to any other private information of the user (e.g., personal details or usage patterns) as well as any of the users in the system, including the relations among users.

The rest of this paper is organized as follows. Section~\ref{sec:related} provides a detailed summary of related work in techniques and tools to help users manage privacy. Section~\ref{sec:approach} explains our approach in detail. Section~\ref{sec:evaluation} provides details on the evaluation setup. Section~\ref{sec:results} evaluates the proposed approach on a widely used data set and demonstrates the benefits of capturing and exploiting uncertainty. Finally, Section~\ref{sec:discussion} concludes our work with pointers to future directions.

\section{Related Work}
\label{sec:related}

The literature on approaches that help users manage their privacy is broad. One of the earlier works is due to Fang and LeFevre \cite{fang2010privacy}, who introduce a wizard software based on an active learning paradigm. The wizard generates a privacy preference model using extracted features from visible data and communities, and also user input such as asking questions. The wizard recommends privacy preferences to users for different information items on their profiles, such as birthday, address, or telephone number. One of their key findings is that a user's social network structure is an important resource when modeling the user's privacy preferences. This idea has been exploited also by Kepez and Yolum \cite{kepez2016learning}, where they propose a machine learning (ML) based model for image privacy prediction. Their framework is based on several attributes about posts such as the sharing time, the location, and the content of the post. They make use of the user's social network to improve prediction. Both of these approaches are important for the privacy prediction task because their approaches use information about the user, her network, and her posts to improve prediction. However, in situations where such external information is not available, there is still a need to make recommendations to the user based only on the content.

Squicciarini {\it et al.} \cite{squicciarini2014privacy} propose Adaptive Privacy Policy Prediction (A3P) system that predicts a privacy policy for images based on the information available for a given user in the context of social networks. A3P needs a user to specify some privacy policies before making a prediction of privacy policies. When recommending a privacy policy for an image, A3P takes into account significant resources for the privacy concept such as actual image content, metadata, and social circle. A3P consists of two main components: A3P-core and A3P-social. When a user uploads an image, the A3P-core classifies the image first based on their contents and then, updates each category into subcategories based on their metadata (if exists). Then, A3P-core either predicts a policy based on the historical behavior or invokes A3P-social. A3P-social finds a representative privacy policies using user's social circle. While the A3P achieves high accuracy, it makes use of information beyond the images themselves and does not attempt to capture the uncertainty in the prediction as we have done here.

Various approaches have exploited using textual and visual features to train classifiers.  An earlier work is by Zerr {\it et al.} \cite{zerr2012privacy}, where they identify that combination of textual and visual features produces the best performance in terms of prediction. However, they do not consider personalization or uncertainty. Tran {\it et al.} \cite{tran2016privacy} propose a privacy framework, called Privacy-CNH that consists of object and convolutional features using a convolutional neural network (CNN) for image privacy detection. Similarly, Squicciarini {\it et al.} \cite{squicciarini2017toward} present a learning model to privacy labels of images for binary privacy labels as private and public as well as multi-class privacy labels such as \textit{Only You} or \textit{Family}. They show that combining scale-invariant feature transformation (SIFT) and tag features perform better than the other two or three combinations such as sentiment, RGB, or facial detection. The results of these approaches have been improved by Tonge and Caragea \cite{tonge2020image}, who tackle the same problem using deep visual semantic and textual features, namely deep tags and user tags. While extracting deep features, they use pre-trained CNN architectures such as AlexNet \cite{krizhevsky2012imagenet}, GoogLeNet \cite{szegedy2015going}, VGG-16 \cite{simonyan2014very}, and ResNet \cite{he2016deep} with Support Vector Machine (SVM) classifiers for the privacy prediction task. Deep tags of images are top \textit{k} predicted object categories that are extracted from pre-trained models. Using user-created tags, they create deep visual features by adding highly correlated tags to visual features extracted from the fully connected layer of the pre-trained models. Their results show that a combination of user tags and deep visual features from ResNet with the top $350$ correlated tags yield the best performance. Moreover, based on their experimental results, fine-tuned networks perform better than learning models trained on the pre-trained features. While their focus has been on classification alone, here, we attempt to take into account both the uncertainty and the personalization associated with making privacy decisions, which is critically important for real-life use cases.

Alternative to approaches that use the image content, some recent approaches have used the tags associated with the content to predict privacy labels of images. 
Squicciarini {\it et al.} \cite{squicciarini2017tag} introduce Tag-To-Protect (T2P) system that automatically recommends privacy policies using the image tags and privacy policies. Their proposed system is useful for both newly uploaded images and cold-start problems when there are very few tags available. One of the prominent results from their experiment is that the prediction accuracy decreases when there is a large set of tags. Since, if the number of tags per image increases, finding a pattern becomes difficult. Kurtan and Yolum \cite{kurtan2021assisting} propose an agent-based approach that predicts the binary privacy labels of images such as private or public using automatically generated content tags. The system keeps track of content being shared using tag tables. The internal tag table stores the data of privacy labels that are collected from images that the user shares herself. The external tag table stores the data collected from the images that the user’s friends have shared with the user. Using metrics inspired by information retrieval, they define metrics to measure how informative a tag is to assess the privacy of an image. Contrary to previous approaches, this system performs well even the personal assistant has access to small data. However, they are not concerned about capturing the uncertainty explicitly or take into account personal risk factors as we have done here.

An alternative set of approaches make use of groups of users, considering various similar aspects among users to make recommendations. Misra and Such~\cite{misra2017pacman} develop PACMAN, a personal assistant that recommends access control decisions. Their approach is based on identifying communities (such as friend networks) from the OSN structure of a user and information about the content, such that users manually select tags to extract information about the content.  Zhong {\it et al.} \cite{zhong2017group} propose Group-Based Personalized Model (GBPM) for an image privacy classification task. Their proposed model learns privacy groups and private content types. Using addition profile information (e.g., gender or age-range), they estimate new users' privacy decisions. They evaluate their proposed model on a randomly selected subset of the PicAlert dataset \cite{zerr2012privacy} by first extending it with by adding demographics and social network usage information. They show that GBPM (with profile information) outperforms several baselines such as SVM approaches.

Fogues {\it et al.} \cite{fogues2017sosharp} present a personal agent, SoSharP that recommends sharing policies in multiuser scenarios. SoSharP uses contextual-based, user-based, preference-based, and group-based features. These features help to provide personalized recommendations in three rounds. SoSharP makes recommendations to each user by using context-based and user-based features in the first round. It moves to the second round if at least one user has not accepted sharing policy. It uses preference-based features in addition to the features used in the first round. In the final round, it makes a recommendation for all users by using group-based features. As a result of the last round, SoSharP recommends manual resolution if most of the users do not agree with the recommendation. Mosca and Such~\cite{mosca2021elvira} also propose an agent, ELVIRA, that for multi-user settings that benefits from recommending individual decisions to each user. 
While we do not consider multi-user settings here, our work can be applied in multi-user settings to recommend privacy labels to each user before a group decision is taken.

Sensoy {\it et al.} \cite{sensoy2021misclassification} propose risk-calibrated evidential deep classifiers to make a better classification by decreasing the costs of misclassified predictions. They reformulate EDL method in order to accomplish this goal. Their experiments show that the proposed Risk EDL method has lower misclassification costs compared to EDL, standard learning with cross-entropy loss, and cost-sensitive learning methods for MNIST, FashionMNIST, and CIFAR10 datasets. They also report that their method is robust for out-of-distribution samples. However, they do not apply their model on privacy as we have done here.

Yu {\it et al.} \cite{yu2018leveraging} propose an algorithm to recommend privacy labels of images in OSN. Their recommendation algorithm takes into account two approaches: an image content sensitiveness and trustworthiness of a user. To train a tree classifier, the algorithm uses feature-based and object-based approaches for the image content sensitiveness and characterization of users' trustworthiness based on social behaviors. Through extensive evaluations over user study and two publicly available datasets (such as PicAlert and MIRFLICKR), they have shown that the proposed algorithm is effective. However, \pure~is both uncertainty-aware and risk-aware model for predicting privacy labels of images. It can achieve high performance without using the profile information of a user, his/her social connections, or organizing different types of image privacy concerns.

Kokciyan and Yolum \cite{kokciyan2020turp} propose an approach, TURP, that manages the trustworthiness of information sources, Internet of Things (IoT) devices, for making context-based privacy decisions. They represent IoT devices and users as software agents. Each agent has a confidence value when it shares information with another agent. In the beginning, each device has the same trust value. These values are updated based on feedback that is given from multiple agents. TURP uses Disjunctive Datalog while reasoning about information collected from multiple agents. It would be interesting to couple TURP with our proposed approach in IoT context so that the privacy decisions are augmented with trust.

Jiao {\it et al.} \cite{jiao2020ieye} design a system, IEye, that provides a personalized and interpretable privacy model. They first extract features from images and use multi-layered semantic graphs for feature representations of images. Then, they learn personalized privacy rule sets from images using the rule-based classification algorithm RIPPER. They compare their methods with SIFT and deep features extracted from pre-trained networks AlexNet, VGG16, and ResNet152. They evaluate the performance of their method IEye on the PicAlert dataset and a small dataset called PPP, which consists of 8744 images of 20 users. IEye has a better accuracy result on the PPP dataset than the baseline approaches. However, the proposed method is not better than deep features for the PicAlert dataset.

Dammu {\it et al.} \cite{dammu2021explainable} develop a system for the image privacy prediction task. Their approach is capable of personalization, explainability, configurability, and customizable privacy labels. The system has four modules such as object detection, location detection, object localization, and explicit content extraction. The decision network aggregates outputs of modules for personalized privacy predictions. This comprehensive approach help make personalized prediction of the image labels. To provide such a personalized system, their approach gets feedback from users for misclassified images. Because of the subjectivity privacy, asking users and using their explanations have an important role. However, it is not clear how this would scale in applications that use large image sets.

Han {\it et al.} \cite{han2021learning} propose a method which uses multi-level and multi-scale deep representations for the image privacy prediction task. First, they obtain these deep representations CNN based model. Then, they propose two feature aggregation models such as Privacy-MSML and Privacy-MLMS based on different aggregation strategies using Bi-LSTM and self-attention. They evaluate the performance of proposed models on a subset of PicAlert dataset. They show that their proposed aggregation models yield better performance by F1-score compared with ResNet-18, CNN-RNN, and concatenated multi-level features. However, they are not concerned with capturing the uncertainty in their predictions as we have done here.

 
\section{PURE: Uncertainty-aware Privacy Assistant}
\label{sec:approach}
\begin{figure}[htb]
    \centering
    \begin{subfigure}[b]{0.44\linewidth}        
        \centering
        \includegraphics[width=\linewidth]{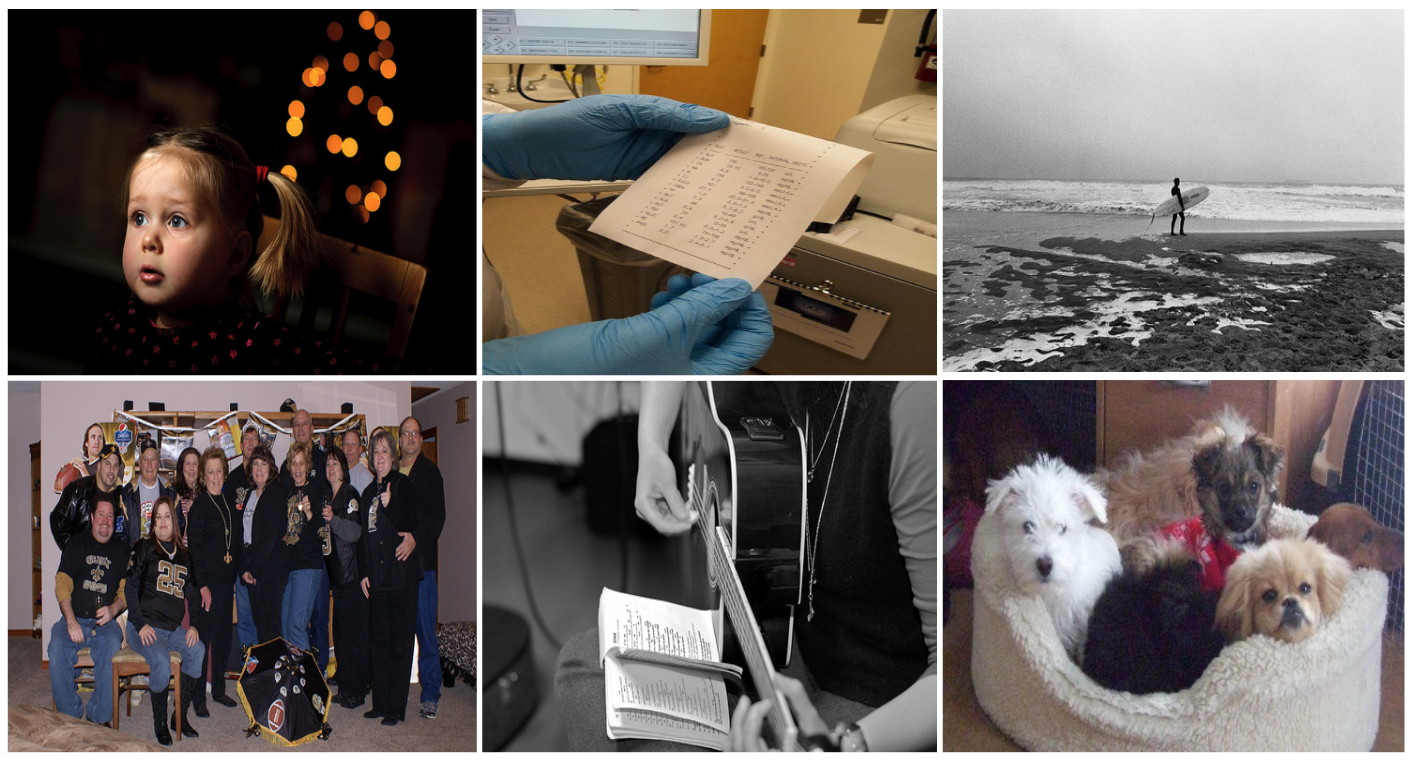}
        \caption{Private images}
        \label{fig:private_class_example}
    \end{subfigure}
    \hspace{0.5in}
    \begin{subfigure}[b]{0.44\linewidth}        
        \centering
        \includegraphics[width=\linewidth]{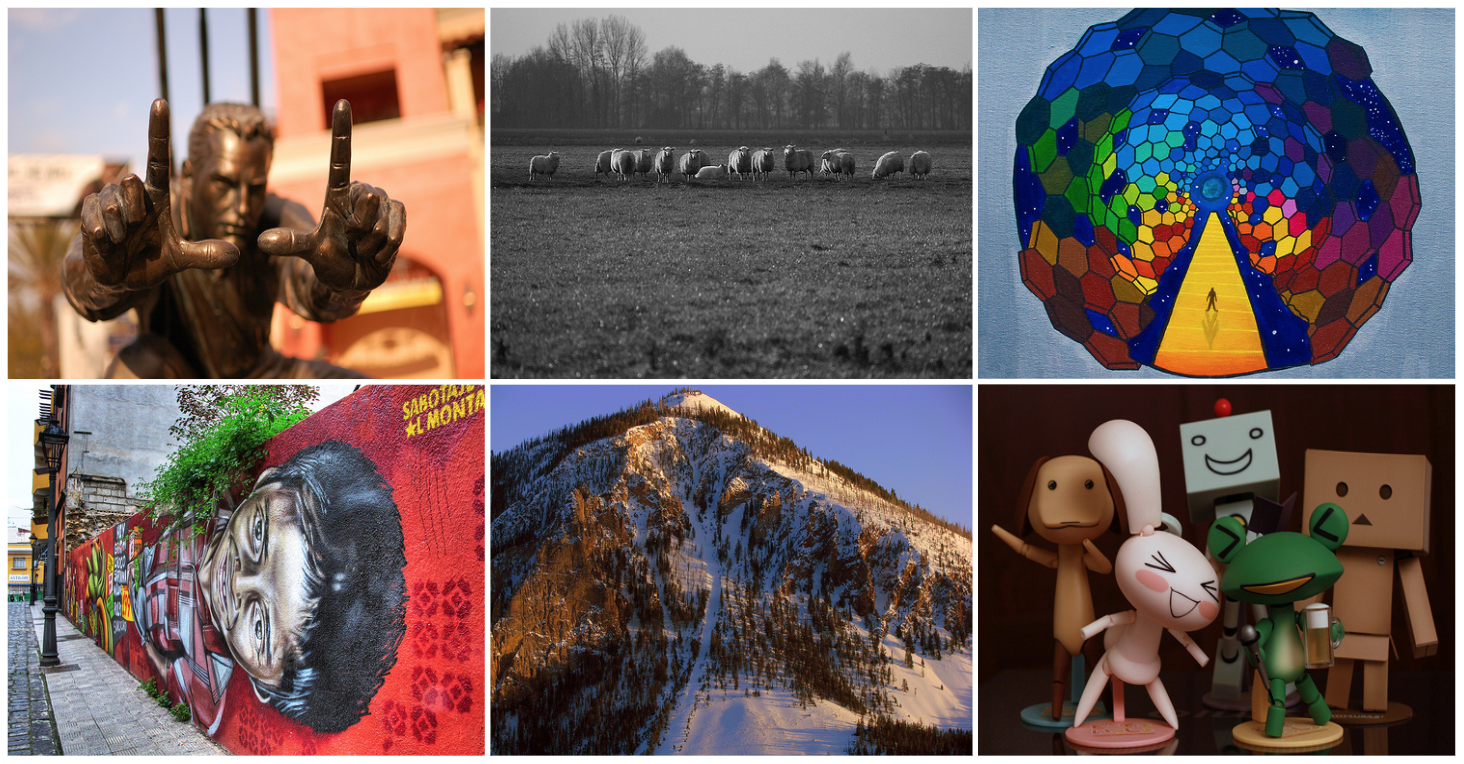}
        \caption{Public images}
        \label{fig:public_class_example}
    \end{subfigure}
    \caption{Examples of images labeled as private and public by the annotators.}
    \label{fig:private_public}
\end{figure}

\begin{figure}[htb]
	\centering
	\includegraphics[scale=0.5]{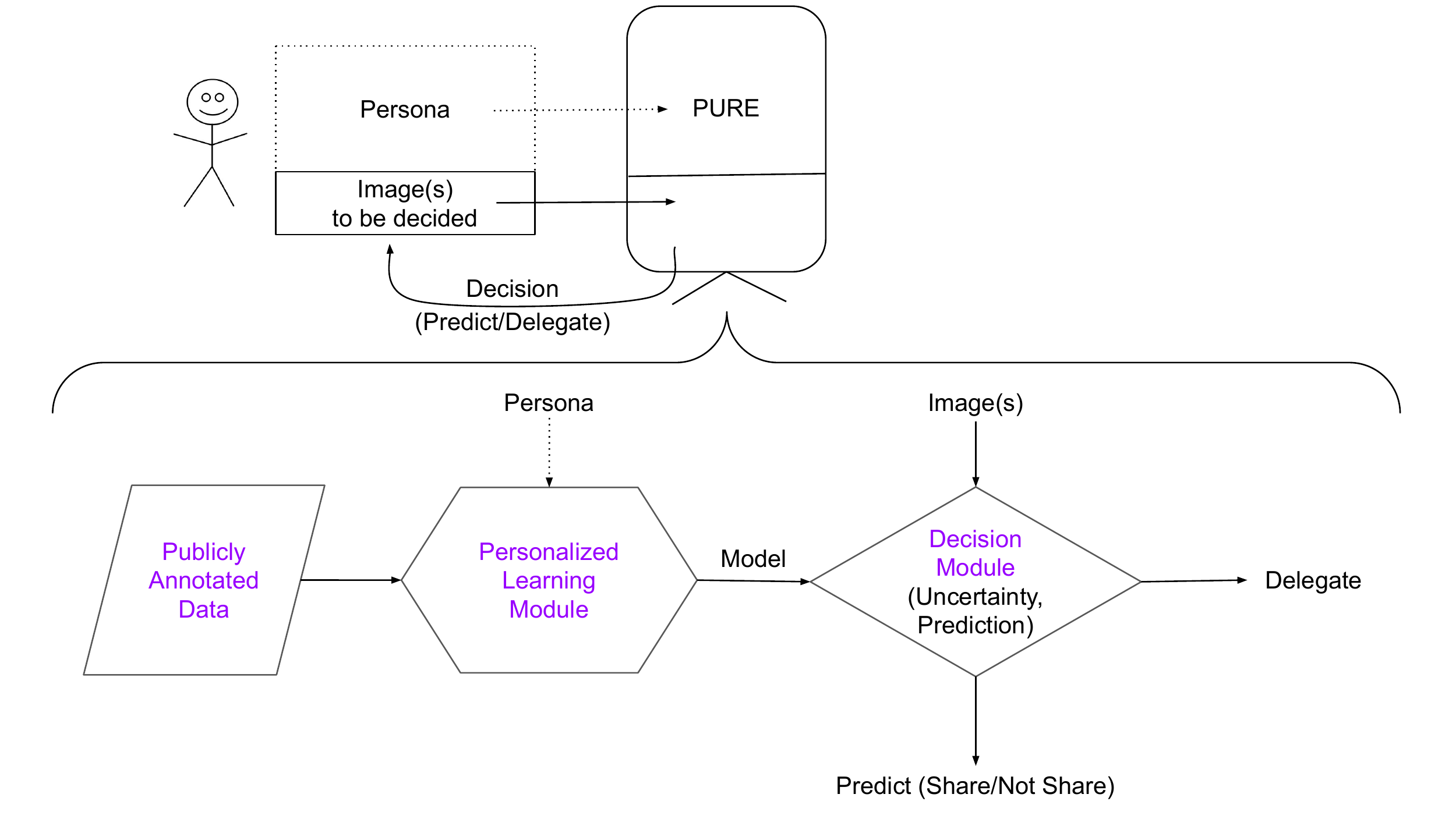}
	\caption{\textit{System Overview Schema:} OSN user has a personal assistant \pure. She can share persona with her personal assistant. First, \pure~has \textit{Publicly Annotated Data} collected from different annotators available in the PicAlert dataset. It learns privacy preferences using visual features in the \textit{Learning Module} and produces \textit{Model}. While learning, the user can share her persona type (i.e., sensitive or non-sensitive) that she can be sensitive about classifying private images as public or not. In this case, the personal assistant is risk-averse.
    Moreover, the user can share personal data that the user herself annotates allowing learning the user's privacy preferences. Then, \pure~makes privacy decisions for its user's content (e.g., image) in the \textit{Decision Module}. While making a prediction for each image, it also generates an uncertainty value for that prediction. To reach a privacy decision, \pure~decides whether to use prediction results (i.e., \textit{share/not share}) or to delegate the decision to the user (i.e., \textit{delegate}) by comparing the uncertainty value with the threshold received from its user.}
	
	\label{fig:system_overview_schema}
\end{figure}

We envision \pure~to work side-by-side with its user when the user is about to share content and help its user make privacy decisions (Figure~\ref{fig:system_overview_schema}).  \pure~uses a learning model that predicts a privacy label of a given image either \textit{private} or \textit{public}. The image is considered to be \textit{private} if it belongs to the private sphere or contains objects that the user cannot share with the world and \textit{public} otherwise. Figure \ref{fig:private_public} shows examples of images as annotated as private and public in the PicAlert dataset by two different annotators. 

\pure~consists of two modules. The main module is the personalized learning module and serves as the core of the personal assistant. The purpose of this module is three-fold. First, using publicly annotated data, it learns to classify images as private or public. Second, it quantifies uncertainties in predictions, such that when it estimates a prediction to be highly uncertain, it can delegate the decision making to the user. Three, it incorporates the user's expectations in privacy, as each user might have different {\it persona} when it comes to how they would like to treat certain factors in the learning. The learning model uses evidential deep learning to realize these goals. The learning module produces a classification model that can label a given image and estimate the uncertainty in the prediction.  Whenever user provides personally labeled data, this module uses that to tune the personal assistant using the images of the user. This is important because privacy is inherently subjective and the publicly annotated dataset that is used for the learning module may not reflect the privacy expectations of the user.  Moreover, due to the subjectivity,~\pure~might assign high uncertainty to some images. By fine-tuning using personal data, we aim to decrease the uncertainty that \pure~might observe with some images. 
The second module is the decision making module. When a user needs to make a privacy decision, this is the module that is invoked. This module obtains a prediction and an uncertainty value from the model. Each user defines for themselves when to let \pure~make a decision and when they would want to be involved. By setting a threshold, a user can choose to decide on the privacy labels when the uncertainty is above the set threshold. Otherwise, the prediction of the model is assigned as the label.

\subsection{Learning Privacy Labels with Uncertainty}
Evidential Deep Learning (EDL) \cite{sensoy2018evidential} is based on Dempster-Shafer theory of evidence \cite{dempster1968generalization} and subjective logic (SL) \cite{josang2016subjective} to quantify uncertainty in classification tasks.
SL expresses degrees of uncertainty through subjective opinions. Each subjective opinion corresponds to a Dirichlet distribution, a conjugate prior for the categorical distribution.
For a binary proposition (e.g., the image $x$ is private), the subjective belief of an agent for the truth of this proposition~\cite{zhang2008detailed, kaplan2014trust} is represented as a binomial opinion, which corresponds to a Beta distribution --- a special form of Dirichlet distribution.
Since privacy classification is a binary classification task, an agent's belief for an image to be private is represented as a binomial subjective opinion.
A binomial subjective opinion for the classification can be represented as a Beta distribution. That is why, in this section, we will introduce EDL using Beta distributions. 
Beta probability density function (pdf) is expressed as:
\begin{equation}
    Beta\left (p| \alpha, \beta  \right ) = \frac{p^{\alpha-1} (1-p)^{\beta-1}}{B\left ( \alpha, \beta   \right )}   
    \label{eq:dirichlet}
\end{equation}
where $B$ is the multivariate beta function \cite{sensoy2018evidential} and $[\alpha, \beta]$ are the parameters of the \textit{Beta} distribution. In Equation \ref{eq:dirichlet}, $p$ is the Bernoulli probability that the binary proposition is true, e.g., the probability that the image $x$ is private.
An agent has a belief ($b$) for the proposition that \textit{the image is private}, a disbelief ($d$) for the same proposition, and an uncertainty ($u$) that represents the inability to classify the image accurately. The agent's uncertainty about an image may be due to the noise in the image or the lack of training data with similar images. We can calculate these quantities as:

\begin{equation*}
\large
    b = \frac{\alpha -1}{\alpha + \beta} \text{,} \quad d = \frac{\beta -1 }{\alpha + \beta},   \quad \text{and} \quad u = \frac{2}{\alpha + \beta},  
\label{eq:uncertainty}
\end{equation*}

where $b, d, u > 0$ and $b+d+u=1$. Furthermore, $\alpha -1$ and $\beta -1$ are called the evidence for and against the proposition: \textit{the image is private}.
Let us note that $u$ is maximized when $\alpha=\beta=1$, corresponding to the uniform Beta distribution.
We can also call them the evidence for the \textit{private} and \textit{public} categories in the classification of the image.

The Beta distribution provides a probability distribution over $p$ --- the probability that the given image is private.
However, in classification tasks, we need a predictive categorical distribution to decide.
For this purpose, we use the expected value of the Beta distribution, which is calculated as follows:
\begin{equation}
    \bar{p} =\int_{0}^{1} p \big( \frac{p^{\alpha-1} (1-p)^{\beta-1}}{B\left ( \alpha, \beta   \right )} \big) dp =\frac{\alpha}{\alpha + \beta}
    \label{eq:expected}
\end{equation}

The aforementioned calculations of belief masses and uncertainty are based on the parameters of the corresponding Beta distribution.
In order to model belief masses and learn Beta distribution parameters, EDL modifies a vanilla neural network for classification by replacing its softmax layer with a non-negative activation function such as \textit{ReLU}, \textit{softplus}, and \textit{exponential} functions.
In our classification problem, we have two categories: \textit{private} and \textit{public}. Given a sample image $x$, we can use any neural network with two logits outputs: $o_0(x)$ and $o_1(x)$, one for each category.
Then, we use the exponential function to calculate evidence for each category as follows:
$e_{pub}(x)=exp(o_0(x))$ and $e_{pri}(x)=exp(o_1(x))$, which represent the evidence for the public and private categories, respectively.
The Beta distribution parameters $\alpha$ and $\beta$ for the classification of the image $x$ are calculated as $\alpha(x) = e_{pri}(x) + 1$ and $\beta(x) = e_{pub}(x) + 1$, respectively. 

Let $y \in \{0, 1\}$ represent the category index of the sample image $x$. 
In standard neural networks for binary classification, the sigmoid function is used to calculate $p(x)=P(y=1|x)$, i.e., the probability that $x$ is from category $y=1$. Then, the binary cross-entropy loss is calculated as follows:
$$
y \log \big( p(x)\big) + (1-y) \log \big(1 - p(x)\big)
$$
There are also other loss functions for classification, such as the Brier score, which is defined as 

\begin{equation}
\label{eq:brier}
   [p(x) - y]^2 + [1-p(x) - (1-y)]^2. 
\end{equation}

The Brier score is a proper scoring function and is frequently used to measure the accuracy of probabilistic predictions. 
Unlike vanilla neural classifiers, we do not predict $p(x)$ directly, so we cannot directly use any of these loss functions. However, we predict its Beta distribution $Beta\big(p(x)|\alpha(x), \beta(x)\big)$; hence, we may calculate the expected loss by integrating out $p(x)$ in the classification loss of our choice. 
We can calculate the expected Brier score for privacy classification as follows:
\begin{equation}
\begin{split}
\mathcal{L}(x,y) = \int_0^1 \big[p(x) - y \big]^2 + \big[1-p(x) - (1-y) \big]^2\\ 
\frac{p(x)^{\alpha(x)-1} (1-p)^{\beta(x)-1}}{B\left ( \alpha(x), \beta(x)   \right )} dp(x)
\end{split}
\end{equation}
which has the following closed-form solution:
\begin{equation}
\begin{split}
    \mathcal{L}(x,y) = [\bar{p}(x) - y]^2 + [1-\bar{p}(x) - (1-y)]^2 +\\
    ~~~~~~~ + 2 \frac{\bar{p}(x) (1-\bar{p}(x))}{\alpha(x) + \beta(x) + 1},
\end{split}
\label{eq:loss_equation}
\end{equation}
where $\bar{p}(x)$ is the expectation of $p(x)$ and calculated as $\alpha(x)/\big(\alpha(x) + \beta(x)\big)$ using Equation \ref{eq:expected}.
%
%
\begin{equation}
\begin{split}
\mathcal{L}(x,y) = \int_0^1 \Big[y \log \big( p(x)\big) + (1-y) \log \big(1 - p(x)\big)\Big]\\ 
\frac{p(x)^{\alpha(x)-1} (1-p)^{\beta(x)-1}}{B\left ( \alpha(x), \beta(x)   \right )} dp(x)
\end{split}
\end{equation}
which has the following closed-form solution:
\begin{equation}
\begin{split}
    \mathcal{L}(x,y) = y\left ( \psi \left ( \alpha(x) + \beta(x) \right ) - \psi\left ( \alpha(x) \right )\right ) +\\
    ~~~~~~~ \left ( 1-y \right )\left ( \psi \left ( \alpha(x) + \beta(x) \right ) - \psi\left ( \beta(x) \right ) \right ),
\end{split}
\label{eq:loss_equation2}
\end{equation}
where $\psi\left ( . \right )$ is the \textit{digamma} function.
%
We also add a regularizing term $\mathcal{R}(x,y)$ to this loss. $\mathcal{R}(x,y)$ is defined as follows:  
\begin{equation} 
\hspace{-0.02in}
\mathcal{R}(x,y) = \lambda_{t} \KL{Beta \big(p(x);\bar{\alpha}, \bar{\beta}\big)}{Beta \big(p(x);1,1 \big)}
\end{equation}
where 
\begin{itemize} 
    \item $t \geq 0$ is the index of the current training epoch,  
    \item $\lambda_{t} = min \left (1.0, t/10  \right )$ is the annealing coefficient,
    \item $\text{KL}[\cdot||\cdot]$ refers to the Kullback-Leibler (KL) divergence,
    \item $\bar{\alpha}=\alpha(x)^{1-y}=(e_{pri}(x)+1)^{1-y}$,
    \item $\bar{\beta}=\beta(x)^y=(e_{pub}(x)+1)^y$,
    \item $Beta \big(p(x);1,1 \big)$ is the uniform Beta distribution.
\end{itemize}
Let us note that $\bar{\alpha}$ and $\bar{\beta}$ do not contain any evidence supporting the true category of the sample image. That is, $\alpha(x)^{1-y}$ becomes $1$ if the image is private ($y=1$) and $\beta(x)^y$ becomes $1$ when the image is public ($y=0$). 
As a result, the KL-divergence term is minimized when the network does not produce any evidence for the wrong category. 
Hence, this regularization term minimizes the evidence generated by the network for the wrong category and increases the predictive uncertainty for the misclassified samples \cite{sensoy2018evidential}.

\begin{figure}[htb]
	\centering
	\includegraphics[scale=0.45]{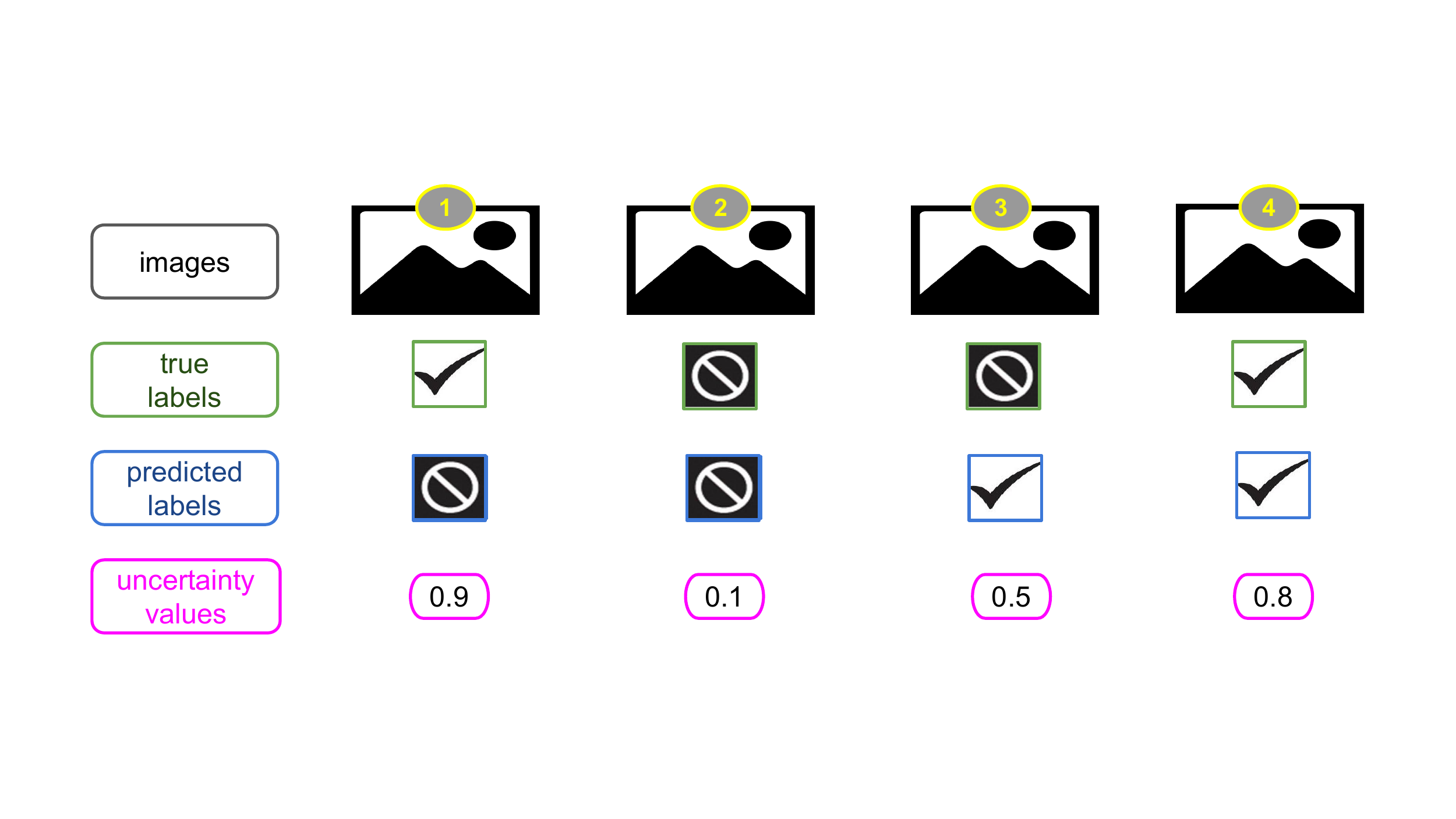}
	\caption{An example for predicting privacy labels of four images and quantifying uncertainty values for each prediction.}
	\label{fig:example_1}
\end{figure}

\bexa\label{ex:predictions} Let's assume that Alice is an OSN user. She has four different images. She needs to decide which images should be shared as public and which should be shared as private. Figure \ref{fig:example_1} represents an example for predicting privacy labels (such as private or public) of her images and quantifying uncertainty values for each prediction. In Figure \ref{fig:example_1}, the first and the fourth images are public, and the other two are private. As shown in Figure \ref{fig:example_1}, \pure~predicts a label for each content as well as an uncertainty value. When producing an answer, it checks its uncertainty value and threshold to decide to answer with its current predicted label or delegate the decision to its user. If the threshold here is $0.7$, it will put forward its predictions for content $2$ and $3$ (as these have uncertainty values $0.1$ and $0.5$, respectively) and delegate content $1$ and $4$ to its user. With this setup, \pure~would have correctly classified image $2$ but image $3$ would have been misclassified. It would have delegated image $1$ that it would have failed on but it would have also been delegated image $4$ to the user that this image is correctly classified.
\eexa

\subsection{Personalizing Privacy} 
Since privacy is inherently subjective, it is important to incorporate personal traits of the user into the decision-making. We consider three aspects of a user that should be factored into the decision making: 1) perception of risk, 2) personal categorization, and 3) preference to be involved.

\mypara{Perception of risk:}
While the personal assistant is making decisions, it is possible that it makes a prediction error. It is possible that for some users misclassifying a private image as public may lead to less desirable consequences than misclassifying a public image as private. For some others, there might not be a difference. 
Furthermore, the cost of different misclassifications may be significantly different for two different users. In order to avoid mistakes that are deemed risky for the user, the system needs to incorporate the risk percention of the user into account. 

Typically, vanilla neural networks do not differentiate this significant difference and consider all mistakes as equal. To overcome this, here we introduce a user-dependent risk matrix, which is an asymmetric non-negative square matrix $R \in [0, \infty)^{2 \times 2}$. 
Each value $R_{ij}$ in $R$ represents the user's cost when the classifier assigns an image from category $i$ to the category $j$. There is no cost for the user for correct classification, hence $R_{ij} \geq R_{ii}=0$.

There may be different ways of incorporating the user's risk of misclassification into the training of evidential classifiers. In this paper, we propose scaling misleading evidence in the KL-divergence term by modifying $\bar{\alpha}$ and $\bar{\beta}$ as follows:
     $$\bar{\alpha}= \big( R_{01} e_{pri}(x) + 1 \big)^{1-y}$$
    $$\bar{\beta}= \big( R_{10} e_{pub}(x) + 1 \big)^y$$
This allows us to increase the KL-divergence further when evidence for high-risk categories are produced. The PURE gets $R$ from its user and can learn how to generate evidence for each category based on the personalized cost of making misclassification.
If the user is sensitive about classifying private images as public, the agent also becomes sensitive and avoids generating evidence for the private category for equivocal and ambiguous images. 

\begin{enumerate}
    \item Scaling misleading evidence in the KL-divergence term by modifying $\bar{\alpha}$ and $\bar{\beta}$ as follows:
     $$\bar{\alpha}= \big( R_{01} e_{pri}(x) + 1 \big)^{1-y}$$
    $$\bar{\beta}= \big( R_{10} e_{pub}(x) + 1 \big)^y$$
    This allows us to increase the KL-divergence further when evidence for high-risk categories are produced.
    \item Regularizing the amount of misleading evidence directly using the risk of misclassification.
    $$
    (1-y) \bar{p}(x) R_{01} e_{pri} + y \big( 1-\bar{p}(x) \big) R_{10} e_{pub},
    $$
    where $\bar{p}(x)$ is the predictive categorical distribution calculated as $\bar{p}(x) = \alpha(x)/(\alpha(x) + \beta(x))$. This term will be added to the aforementioned loss: $\mathcal{L}(x,y) + \mathcal{R}(x,y)$.
\end{enumerate}

\begin{figure}[htb]
	\centering
	\includegraphics[scale=0.45]{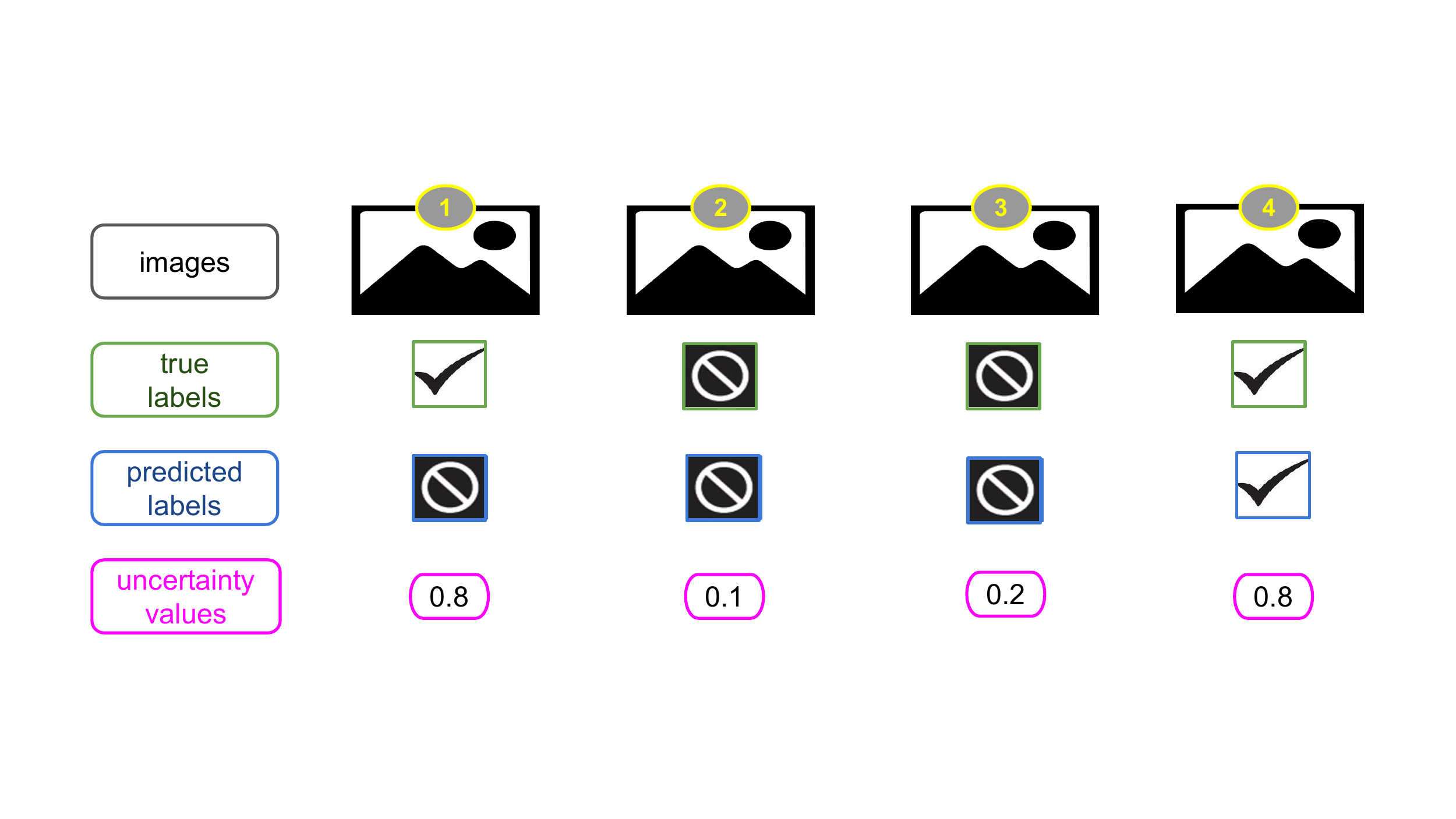}
	\caption{An example for predicting privacy labels of four images for a sensitive user and uncertainty values for each prediction.}
	\label{fig:example_2}
\end{figure}

\bexa\label{ex:persona} If for a user, there is no difference between the misclassification of private and public images, then \pure~makes predictions as shown in Figure~\ref{fig:example_1}. 
On the other hand, assume that Alice is more sensitive about classifying a private image as public. By reflecting this in the $R_{ij}$ score, \pure~predicts privacy labels of images and quantifies uncertainties for each prediction as shown in Figure~\ref{fig:example_2}. Notice that the uncertainty values, as well as the predicted labels, have changed compared to Figure~\ref{fig:example_1}. With the uncertainty threshold still set to $0.7$, \pure~will delegate the same set of images ($1$ and $4$) and answer images $2$ and $3$. Image $3$ has been correctly classified this time. This is a by-product of the fact that \pure~chooses to classify more images as private to avoid the potential risk associated with classifying private images as public. 
\eexa

\mypara{Personal Categorization:}
Another aspect of personalization is to understand what images a particular user finds private or public. One way of understanding this is to ask the user about privacy preferences. However, there is long standing evidence that users are not good at articulating what they find private. Moreover, their actions are not always in line with what they claim to be private. Thus, a better way of understanding what is private for a user is to utilize personal data: images that are labeled by the user herself.

\pure~makes use of this to fine tune the model it generates. After \pure~is trained on publicly annotated data, the user's own labeled data is to adjust the uncertainties in the model. An important contribution of this would be that the uncertainty in certain images drop such that model is more certain of its prediction.

\begin{figure}[htb]
	\centering
	\includegraphics[scale=0.45]{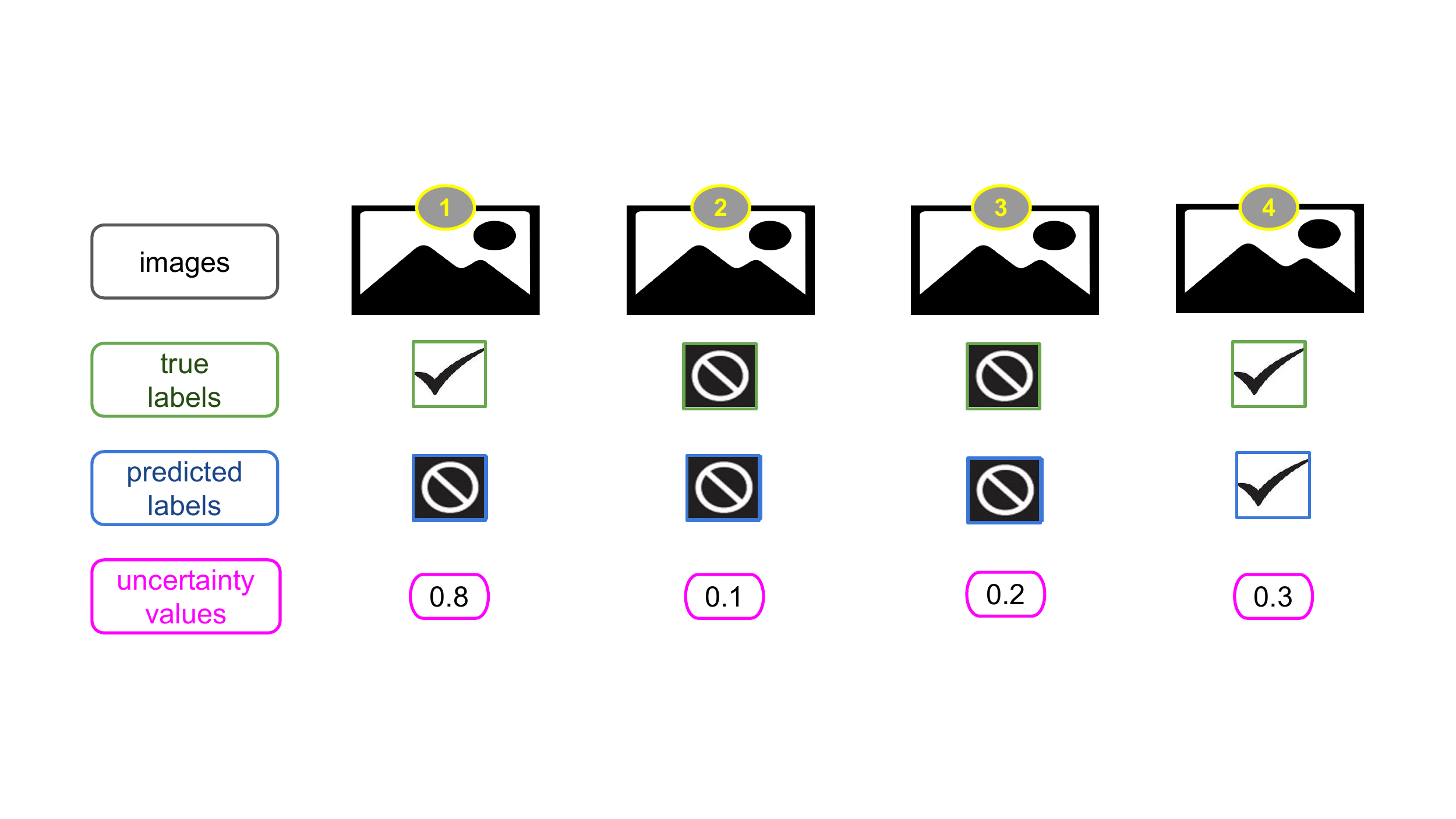}
	\caption{An example for predicting privacy labels of four images by fine-tuning using personal data and uncertainty values for each prediction.}
	\label{fig:example_3}
\end{figure}

\bexa\label{ex:personalization} If \pure~uses publicly available data and if Alice is a sensitive user about classifying a private image as public, \pure~makes predictions in Figure \ref{fig:example_1} and \ref{fig:example_2}, respectively. Moreover, if Alice shares her personal data that has been annotated by her, \pure~will predict privacy labels and uncertainty values for each prediction as shown in Figure \ref{fig:example_3}. If uncertainty threshold is still $0.7$, \pure~will delegate content $1$ to Alice and correctly classify content $2, 3$, and $4$ (as these have uncertainty values $0.1, 0.2$, and $0.3$, respectively). Since content $4$ would have been correctly predicted with a lower uncertainty value, it would have not delegated to the user. So, all the classifications would be correct this time and \pure~would ask its user less.

\eexa
\mypara{Preference to be involved}
The final part of personalization is to understand how much a user wants to be involved in the decision making. Recent work in HCI show that~\cite{colnago-2020} while some users are happy to have privacy decisions taken by their privacy assistants on their behalf, some users would rather be in the loop. Moreover, this is not always a binary decision in the sense that with some decisions the user might want to be involved while with other she might not. We capture this preference to be involved in the decision making process explicitly using a threshold value $\theta$.  Whenever \pure~is asked to label an image, in addition to a prediction, \pure~also provides a level of uncertainty. When \pure~has an uncertainty above $\theta$, it delegates the decision making back to the user. Since $\theta$ can be configured by the user herself, it enable the user to select a level of involvement, where $\theta=1$ would mean letting \pure~do all the decisions, where $\theta=0$ would mean overseeing all the decisions. During our experiments, we discuss having $\theta=0.7$ as a working setting to capture user involvement only when \pure~has high uncertainty.

\section{Evaluation}
\label{sec:evaluation}
We evaluate the performance of \pure~in terms of its contribution to preserving privacy. Specifically, we aim to answer the following research questions:
\begin{description} 
    \item[RQ1]\label{rq-uncertainty} Does \pure~capture the privacy ambiguity through its modeling of uncertainty and by delegating ambiguous cases to the user, can \pure~increase its privacy prediction accuracy?
    \item[RQ2] Does \pure~capture uncertainty adequately and outperform existing models that capture uncertainty? 
    \item[RQ3] Can \pure~enable personalization of privacy by incorporating privacy risks and personal data of the user so that the accuracy is improved for the user while the number of delegated decision decrease?
\end{description}

It is important to be able to answer RQ1 affirmatively because capturing the ambiguity is the key for \pure~to choose when to consult its user. Ideally, uncertain images should be delegated to the user for a decision, and certain images should be answered by \pure. As a result, if we consider only the certain images that \pure~makes a prediction on, we would expect to obtain a higher accuracy than the overall accuracy. RQ2 investigates the dynamics between uncertainty and making prediction errors and questions whether alternative formulations of uncertainty such as an SNN or well-known uncertainty quantification methods such as MC dropout and Deep Ensemble would suffice. Finally, RQ3 explores if and to what extent personalization of \pure~helps users, either in terms of the accuracy they obtain or the number of images they have to decide.

\subsection{Dataset}

To evaluate our work, we selected a balanced subset of the PicAlert dataset \cite{zerr2012privacy}. The PicAlert is a well-known benchmark dataset for the privacy prediction problem for images that contains Flickr images that are labeled as \textit{public} or \textit{private} by external viewers. These images are the most recently uploaded images for four months in $2010$ and labeled by $81$ users between $10$ and $59$ years of age with varied backgrounds. 
$17\%$ of the images in this dataset have conflicting labels from annotators. We consider an image as public if all the annotators have annotated it as public and private if at least one annotator has annotated it as private. The subset we work with contains $32K$ samples that are labeled as \textit{public} and \textit{private}. It is split into \textit{Train} and \textit{Test} sets of $27K$ and $5K$ samples, respectively. 

While the previous research aims at increasing the accuracy of the privacy prediction, additionally we focus on how to quantify the uncertainty in these predictions and exploit it to improve the user's privacy in the face of automated decisions. 

\subsection{Metrics}
We evaluate the performance of our approach using two main metrics: (i) success of the model in terms of the standard metrics such as \textit{Accuracy}, \textit{F1-score}, \textit{Precision}, and \textit{Recall}; and (ii) ability of the model to quantify its predictive uncertainty, which allows the improvement of the success metrics in (i) if quantified correctly and accurately.

We first evaluate our approach without considering personalization; hence the generated evidence is not weighted based on the perceived privacy risk of the user. Then, we extend our evaluations with the personalized risk matrices to see how our model adapts itself for users with different misclassification costs. We also extend evaluations with personal data which is annotated by a user to observe how \pure~adapts and then asks less to its user. 
To evaluate the quality of the uncertainty estimates, we calculate the accuracy of the model only on the test samples for which the model's uncertainty is less than a given uncertainty threshold between $0$ and $1$.
When the uncertainty threshold is $1$, all test samples are considered in computing the accuracy (and other metrics like precision and recall); however, when the threshold is reduced to $0.5$, predictions with uncertainty less than $0.5$ are considered for the calculation.

\subsection{Evaluation Setting} 
We use models which are pre-trained on the ImageNet to extract features from images. We compare three popular deep architectures of convolutional neural networks: ResNet50 \cite{he2016deep}, InceptionV3 \cite{szegedy2016rethinking}, and VGG16 \cite{simonyan2014very} in terms of their performance. 
Table \ref{tab:comparison_pretrained} shows the results of the comparison (\textit{Accuracy}, \textit{F1-score}, \textit{Precision}, and \textit{Recall}) of \pure~using ResNet50, InceptionV3, and VGG16. ResNet50 and InceptionV3 pre-trained models yield better-performing models as compared to VGG16. ResNet50 avoids the network from the vanishing gradient problem. It has Batch Normalization layers that mitigate Internal Covariate Shift. Because it enables more efficient training and performs well, while having fewer layers and parameters for other models with similar performance, we choose \textit{ResNet50} as our underlying architecture.

\begin{table}[htb]
\centering
\begin{tabular}{|c|c|c|c|c|}
 \hline
\textbf{} & \textbf{Accuracy} & \textbf{F1} & \textbf{Precision} & \textbf{Recall} \\ \hline
\textbf{ResNet50}         & 0.89              & 0.89        & 0.89               & 0.89            \\ \hline
\textbf{InceptionV3}        & 0.89              & 0.89        & 0.89               & 0.89            \\ \hline
\textbf{VGG16}        & 0.73              & 0.72        & 0.8               & 0.73            \\ \hline

\end{tabular}
\\[1ex]
\vspace{0.05in}
\caption{Performance of \pure~using different pre-trained models ResNet50, InceptionV3, and VGG16. \label{tab:comparison_pretrained}} 
\end{table}

We use the \textit{ResNet50} architecture as our base neural network and replace its last layer (logits layer) with a densely connected layer with two outputs --- one for each class (private and public). This architecture has $50$ layers with residual connections.
We implement our model using Tensorflow and initialize the network layers from the \textit{ResNet50} model and train 10 epochs on the PicAlert dataset using Adam optimizer with a decaying learning rate initialized as $1e-5$. The \textit{ResNet50} accepts images with dimensions $(224 \times 224 \times 3)$, so we resize the images to these dimensions.

\section{Experimental Results}
\label{sec:results}
We perform the following experiments with the mentioned dataset to answer our research questions.

\mypara{Performance of \pure:} We start with examining the accuracy of \pure, where we configure \pure~to provide a label no matter what the uncertainty is. We experiment with using the entire available training data as well as compare it to cases where the training data is smaller.

\begin{table}[htb]
\centering
\begin{tabular}{|c|cccc|}
\hline
\textbf{}   &       \multicolumn{4}{c|}{\textbf{Overall}}                        \\ \hline
\textbf{\begin{tabular}[c]{@{}c@{}}Usage \\ \%\end{tabular}} &
  \multicolumn{1}{c|}{\textbf{Accuracy}} &
  \multicolumn{1}{c|}{\textbf{F1}} &
  \multicolumn{1}{c|}{\textbf{Precision}} &
  \textbf{Recall} \\ \hline
\textbf{100}  & \multicolumn{1}{c|}{0.89} & \multicolumn{1}{c|}{0.89} & \multicolumn{1}{c|}{0.89} & 0.89 \\ \hline
\textbf{75} & \multicolumn{1}{c|}{0.88} & \multicolumn{1}{c|}{0.89} & \multicolumn{1}{c|}{0.88} & 0.88 \\ \hline
\textbf{50} & \multicolumn{1}{c|}{0.88} & \multicolumn{1}{c|}{0.88} & \multicolumn{1}{c|}{0.88} & 0.88 \\ \hline
\textbf{25} & \multicolumn{1}{c|}{0.87} & \multicolumn{1}{c|}{0.87} & \multicolumn{1}{c|}{0.87} & 0.87 \\ \hline
\textbf{10} & \multicolumn{1}{c|}{0.79}  & \multicolumn{1}{c|}{0.78} & \multicolumn{1}{c|}{0.82} & 0.79  \\ \hline
\textbf{5} & \multicolumn{1}{c|}{0.66}  & \multicolumn{1}{c|}{0.62} & \multicolumn{1}{c|}{0.76} & 0.66  \\ \hline
\textbf{1} & \multicolumn{1}{c|}{0.55} & \multicolumn{1}{c|}{0.44} & \multicolumn{1}{c|}{0.69} & 0.55 \\ \hline
\end{tabular}
\\[1ex]
\vspace{0.05in}
\caption{Overall results for \pure~as training samples are reduced.
\label{tab:overall_training_sample_rejection}}
\end{table}

Table \ref{tab:overall_training_sample_rejection} shows the overall performance of \pure. For instance, when we use all data while training, \pure~obtains an accuracy of $0.89$. \pure~obtains an accuracy of $0.87$ for $25\%$ of training data. On the other hand, if \pure~is trained on only $1\%$ of data, the accuracy decreases to $0.55$. Table \ref{tab:class_based_training_sample_rejection} shows the performances of \pure~for the private and public classes while using with different amount of data. For instance, \pure~achieves F1-score of $0.89$ for each class when \pure~uses all images in the training dataset. While training with $1\%$, \pure~exhibits poor performance in terms of F1-score, especially for the private class. If a user has only $10\%$ of training data, \pure~obtains F1-score of $0.75$ and $0.81$ for the private and public class, respectively. This is promising because it shows that even when there is limited training data, \pure~can be useful. 

\begin{table}[htb]
\centering
\begin{tabular}{|c|ccc|ccc|}
\hline
\textbf{}   & \multicolumn{3}{c|}{\textbf{Private}}                        & \multicolumn{3}{c|}{\textbf{Public}}                         \\ \hline
\textbf{\begin{tabular}[c]{@{}c@{}}Usage \\ \%\end{tabular}} &
  \multicolumn{1}{c|}{\textbf{F1}} &
  \multicolumn{1}{c|}{\textbf{Precision}} &
  \textbf{Recall} &
  \multicolumn{1}{c|}{\textbf{F1}} &
  \multicolumn{1}{c|}{\textbf{Precision}} &
  \textbf{Recall} \\ \hline
\textbf{100}  & \multicolumn{1}{c|}{0.89} & \multicolumn{1}{c|}{0.91} & 0.87 & \multicolumn{1}{c|}{0.89} & \multicolumn{1}{c|}{0.87} & 0.92 \\ \hline
\textbf{75} & \multicolumn{1}{c|}{0.88} & \multicolumn{1}{c|}{0.91} & 0.87 & \multicolumn{1}{c|}{0.89} & \multicolumn{1}{c|}{0.86} & 0.92 \\ \hline
\textbf{50} & \multicolumn{1}{c|}{0.88} & \multicolumn{1}{c|}{0.92} & 0.84 & \multicolumn{1}{c|}{0.89} & \multicolumn{1}{c|}{0.85} & 0.92 \\ \hline
\textbf{25} & \multicolumn{1}{c|}{0.86} & \multicolumn{1}{c|}{0.91} & 0.81 & \multicolumn{1}{c|}{0.87} & \multicolumn{1}{c|}{0.82} & 0.92 \\ \hline
\textbf{10} & \multicolumn{1}{c|}{0.75}  & \multicolumn{1}{c|}{0.92} & 0.63 & \multicolumn{1}{c|}{0.81} & \multicolumn{1}{c|}{0.72} & 0.95 \\ \hline
\textbf{5} & \multicolumn{1}{c|}{0.49}  & \multicolumn{1}{c|}{0.94} & 0.34 & \multicolumn{1}{c|}{0.74} & \multicolumn{1}{c|}{0.6} & 0.98 \\ \hline
\textbf{1} & \multicolumn{1}{c|}{0.19} & \multicolumn{1}{c|}{0.85} & 0.11 & \multicolumn{1}{c|}{0.68} & \multicolumn{1}{c|}{0.52} & 0.98 \\ \hline
\end{tabular}
\\[1ex]
\vspace{0.05in}
\caption{Results for the private and public classes of \pure~at different training sample rates.
\label{tab:class_based_training_sample_rejection}}
\end{table}

\begin{table}[]
\centering
\begin{tabular}{|c|c|c|c|c|}
\hline
                      & \multicolumn{4}{c|}{\textbf{Overall}}                                  \\ \hline
\textbf{\begin{tabular}[c]{@{}c@{}}Delegation \\ \%\end{tabular}} & \textbf{Accuracy} & \textbf{F1} & \textbf{Precision} & \textbf{Recall} \\ \hline
\textbf{0 }         & 0.89              & 0.89        & 0.89               & 0.90            \\ \hline
\textbf{10 }        & 0.92              & 0.92        & 0.92               & 0.92            \\ \hline
\textbf{25 }        & 0.95              & 0.95        & 0.95               & 0.95            \\ \hline
\textbf{50 }        & 0.97              & 0.97        & 0.97               & 0.97            \\ \hline
\textbf{75 }        & 0.99              & 0.99        & 0.99               & 0.99            \\ \hline
\end{tabular}
\\[1ex]
\vspace{0.05in}
\caption{Overall results for \pure~at prediction delegation rates $0\%, 10\%, 25\%, 50\%$, and $75\%$ based on uncertainty. \label{tab:edl_uncertainty_result_all}} 
\end{table}

Recall that an important aspect of \pure~is that it can calculate uncertainty. Next, we look at the relation between uncertainty and accuracy to capture if \pure~can represent uncertainty correctly. We have set up \pure~so that it would delegate to its user when it is uncertain and thus is likely to make a mistake. Hence, ideally, when \pure~delegates to its user, we would expect an improvement in the accuracy of the remaining items.

Table \ref{tab:edl_uncertainty_result_all} shows the overall performance of \pure~with respect to different percentages of delegated predictions. When we do not delegate any predictions, \pure~obtains an accuracy of $89\%$ (as was shown in Table~\ref{tab:overall_training_sample_rejection}). When we delegate only $25\%$ of the most uncertain predictions, the results based on all performance metrics improve remarkably, e.g., the accuracy, recall, and precision increase to $0.95$. Similarly, when we delegate $75\%$ of the most uncertain predictions, \pure~achieves the highest performance of $0.99$ in terms of all metrics. Thus, we observe that the delegated images are actually the ones that \pure~would have made a mistake in.

An important question is whether the same upward trend holds for both private and public class. Table \ref{tab:edl_uncertainty_result_private_public} shows the performance of \pure~for each class. For instance, when \pure~does not delegate any predictions, it obtains F1-score of $89\%$ for both private and public classes. When \pure~delegates only $25\%$ of the most uncertain predictions, \pure~improves F1-scores to $94\%$ and $95\%$ for the private and public classes, respectively. When it delegates $75\%$ of the most uncertain predictions to its user, \pure~yields the best performance with $0.99$ F1-scores for both private and public classes, respectively. By increasing the number of the delegated predictions, the performance of the model can be improved for each class significantly.
\begin{table}[htb]
\centering
\begin{tabular}{|c|c|c|c|c|c|c|}
\hline
                      & \multicolumn{3}{c|}{\textbf{Private}}              & \multicolumn{3}{c|}{\textbf{Public}}               \\ \hline
\textbf{\begin{tabular}[c]{@{}c@{}}Delegation \\ \%\end{tabular}}  & \textbf{F1} & \textbf{Precision} & \textbf{Recall} & \textbf{F1} & \textbf{Precision} & \textbf{Recall} \\ \hline
\textbf{0 }         & 0.89        & 0.91               & 0.87            & 0.89        & 0.87               & 0.92            \\ \hline
\textbf{10}        & 0.91        & 0.94               & 0.89            & 0.92        & 0.90               & 0.94            \\ \hline
\textbf{25}        & 0.94        & 0.96               & 0.92            & 0.95        & 0.94               & 0.97            \\ \hline
\textbf{50}        & 0.97        & 0.99               & 0.95            & 0.98        & 0.96               & 0.99            \\ \hline
\textbf{75}        & 0.99        & 0.99               & 0.98            & 0.99        & 0.98               & 0.99            \\ \hline
\end{tabular}\\[1ex]
\vspace{0.05in}
\caption{Results for the private and public classes of \pure~at various prediction delegation rates ($0\%, 10\%, 25\%, 50\%$, and $75\%$) based on uncertainty.  \label{tab:edl_uncertainty_result_private_public}}
\end{table}

Another dimension to understand the link between uncertainty and making errors is to analyze what fraction of wrong predictions fall under different uncertainty rates. Figures~\ref{fig:fail_success_private_pure} and \ref{fig:fail_success_public_pure} present the uncertainty histogram for the failed and successful privacy prediction of \pure, separately for the private and public classes. The failed and successful predictions in the uncertainty ranges of each class are shown as a percentage among themselves. We observe that failed predictions have higher uncertainty in general while successful predictions are more confident. This indicates that \pure~is aware of its own ignorance and possible failures through its predictive uncertainty. When the most uncertain predictions are eliminated, its accuracy improve drastically. 
\begin{minipage}[t]{0.4\textwidth}
\centering
\begin{tikzpicture}
  \node (img1) {\includegraphics[height=1.6in]{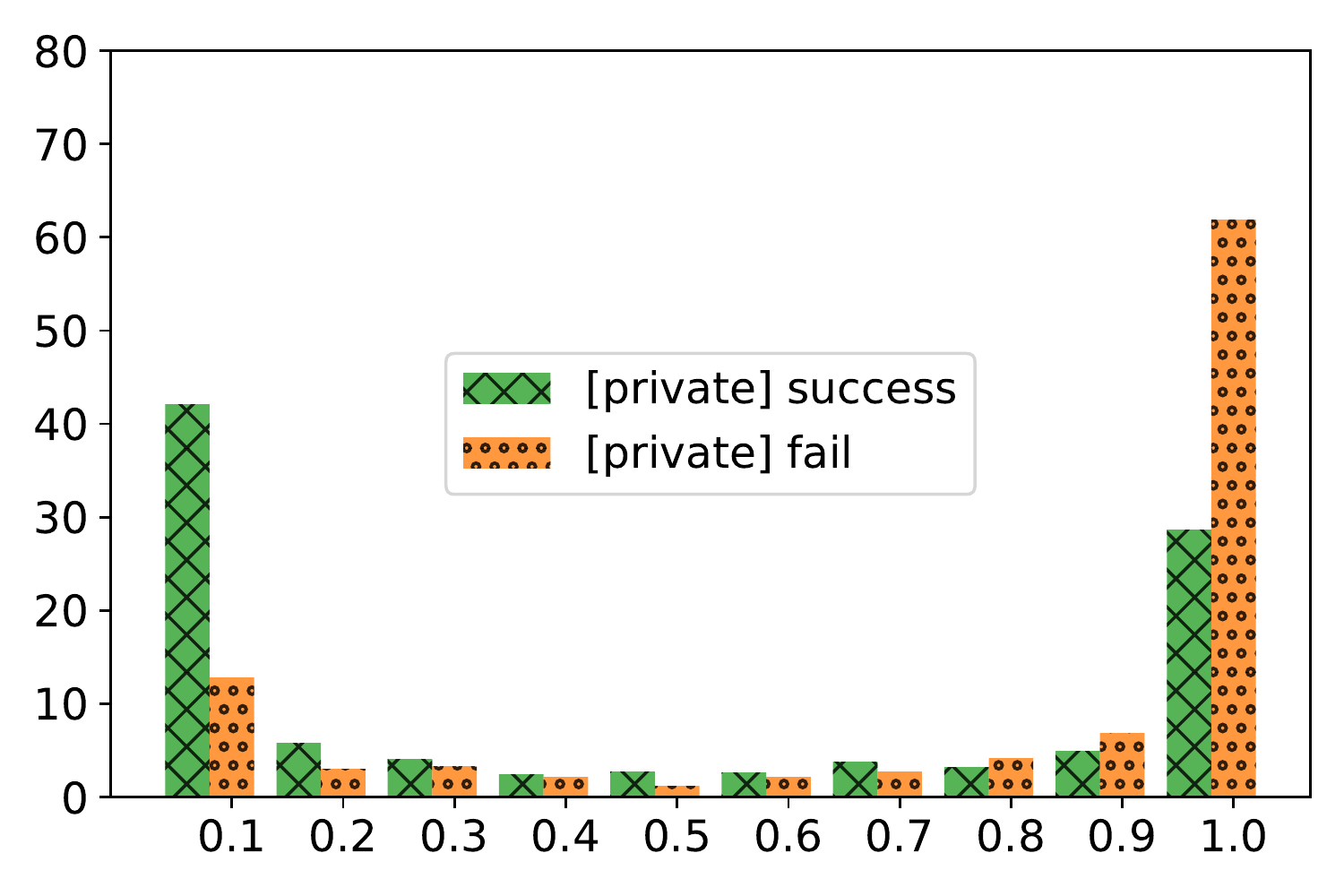}};
  \node[below=of img1, node distance=0cm, yshift=1.2cm] {Uncertainty};
  \node[left=of img1, node distance=0cm, rotate=90, anchor=center,yshift=-1cm] {Percentage of Samples};
 \end{tikzpicture}
  	\captionof{figure}{Uncertainty distribution for the private category.}
  	\label{fig:fail_success_private_pure}
\end{minipage}%
  \hspace{0.5in}
\begin{minipage}[t]{0.4\textwidth}
\centering
\begin{tikzpicture}
  \node (img2) {\includegraphics[height=1.6in]{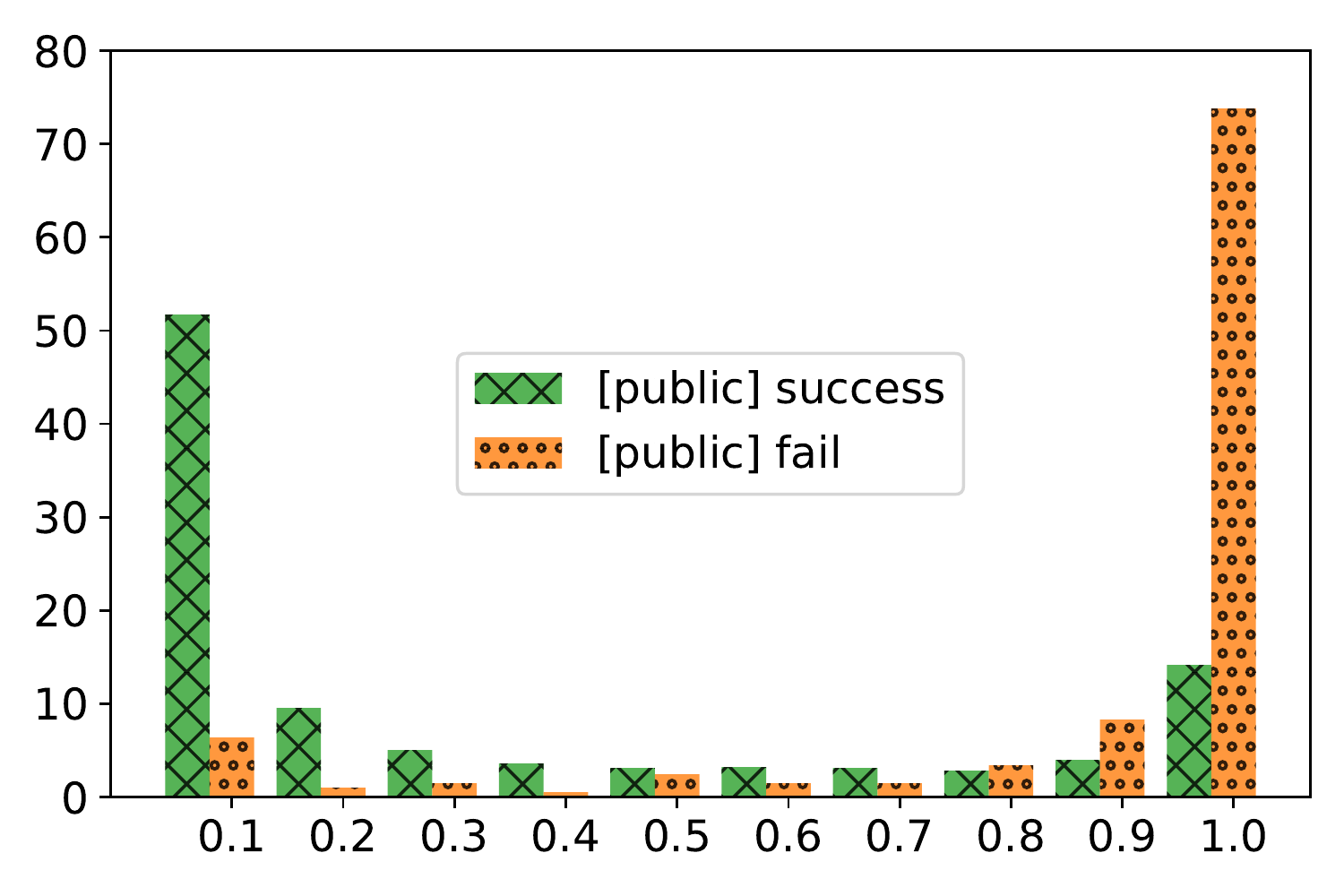}};
  \node[below=of img2, node distance=0cm, yshift=1.2cm] {Uncertainty};
  \node[left=of img2, node distance=0cm, rotate=90, anchor=center,yshift=-1cm] {Percentage of Samples};
\end{tikzpicture}
	\captionof{figure}{Uncertainty distribution for the public category.}
	\label{fig:fail_success_public_pure}
\end{minipage}
\\[1ex]

\begin{figure}[htb]
\centering
\begin{tikzpicture}
  \node (img1)  {\includegraphics[scale=0.75]{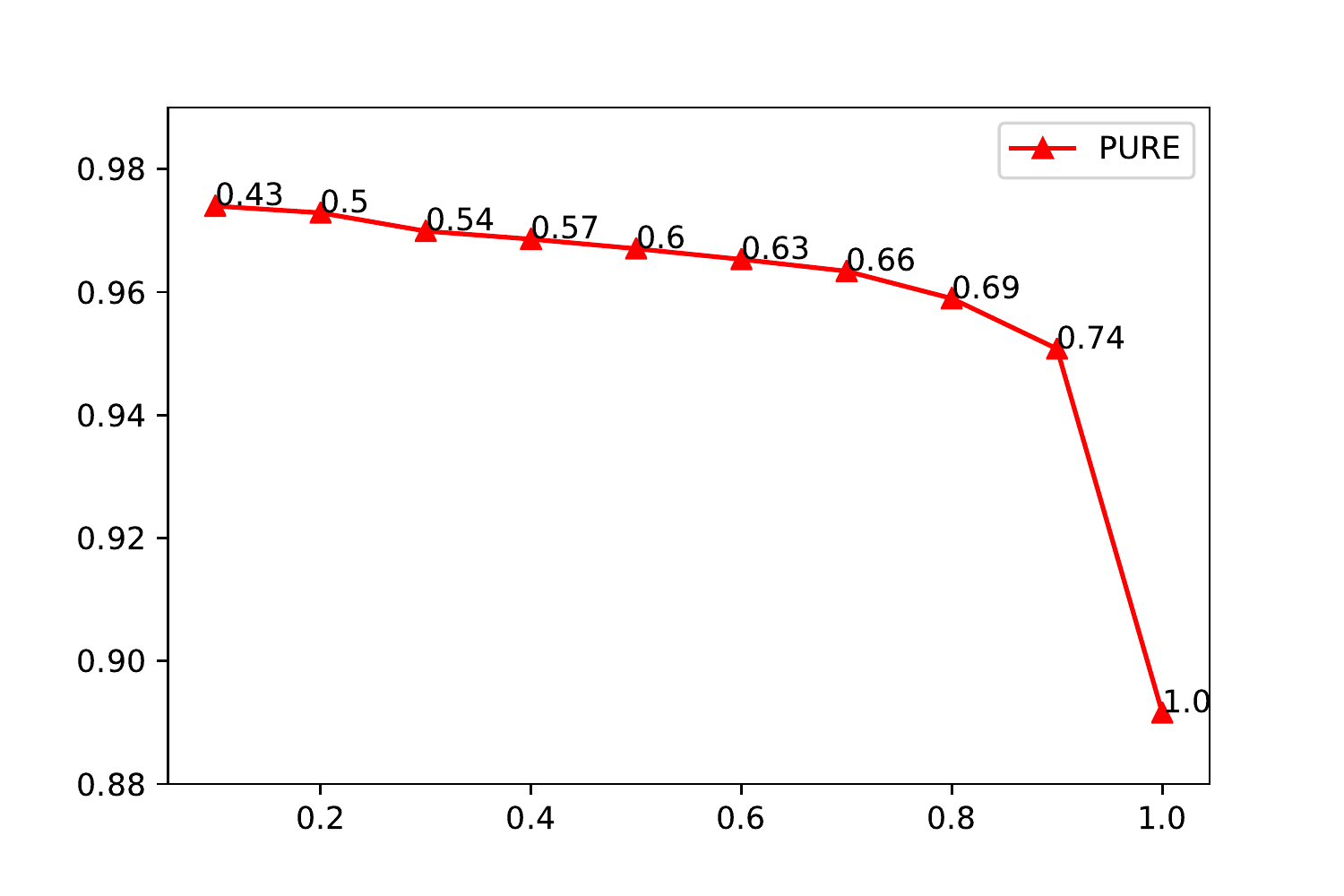}};
  \node[below=of img1, node distance=0cm, yshift=1.5cm] {Uncertainty Threshold};
  \node[left=of img1, node distance=0cm, rotate=90, anchor=center,yshift=-1.5cm] {Accuracy};
\end{tikzpicture}
	\caption{The change of accuracy with respect to the uncertainty threshold.}
	\label{fig:unc_vs_acc}
\end{figure}

Similarly, Figure \ref{fig:unc_vs_acc} plots uncertainty against accuracy for \pure. The numbers on the particular points denote the ratio of test samples that are decided by \pure; the remaining samples are delegated back to the user because of the high uncertainty.  The case when uncertainty is set to $1$ is analogous forcing \pure~to make all the privacy decisions, without delegating any case to the user. The accuracy of \pure~at this stage is $89\%$. This is on par with the existing models in the literature that use the same dataset to predict privacy labels~\cite{tonge2020image}. The more interesting cases are the ones where the uncertainty is high so that the \pure~decides that there is too much uncertainty to answer and delegates them to the user.  For example, for uncertainty threshold $0.4$, $57\%$ of the test samples can be decided by \pure, leading to an accuracy around $0.97$.  For uncertainty threshold $0.8$, $69\%$ of the test samples can be decided with \pure, leading to an accuracy around $0.95$. This shows, as RQ1 asks, that \pure~can capture the privacy ambiguity and it can delegate such cases to its user.

\mypara{Comparison with Alternative Networks:}
The \pure~calculates the uncertainty of its predictions and exploits it to refrain from making wrong privacy decisions for its users. In order to understand the effect of \pure~in its calculations of uncertainty, we compare it to alternative predictive uncertainty models; Monte Carlo (MC) dropout \cite{gal2016dropout} and Deep Ensemble \cite{lakshminarayanan2017simple}, which depend on the class probabilities predicted by the neural network as well as a regular Standard Neural Network (SNN). We implement a SNN, MC dropout, and Deep Ensemble with two softmax outputs using the same \textit{ResNet50} architecture with \pure. In order to measure the uncertainty of standard deep classifiers, entropy of their predictions has been used after normalising it to have an uncertainty value between $0$ and $1$~\cite{sensoy2018evidential,hullermeier2019aleatoric}. 
To have a meaningful comparison for uncertainty quantification, we use the normalized entropy as a proxy for the uncertainty for the \pure~and SNN, MC dropout, Deep Ensemble models.
For \pure, we use the expected probabilities defined in Equation \ref{eq:expected} to calculate the entropy. The entropy for the class probabilities $p$ and $(1-p)$ is calculated as $-[p \log p + (1-p) \log (1-p)]$; then normalized by dividing to $\log 2$, which is the maximum entropy for the binary classification. 
Gal and Ghahramani propose MC dropout method that represents model uncertainty using dropout in neural networks at test time. We add dropout layers after each non-linearities and set the dropout rate as $0.05$. \footnote{$0.05$ yields the best performance among $\left \{ 0.01, 0.1, 0.25, 0.5\right \}$.} We train a model, obtain the desired number of different predictions and take the average of $5$ predictions for each class. Lakshminarayanan {\it et al.} propose ensemble based method, called Deep Ensemble that quantifies predictive uncertainty. We use Brier score (Equation \ref{eq:brier}) as a proper scoring rule as the training criterion, train $5$ models with the same architecture, and take the average of predictions.

\begin{figure}[htb]
\centering
\begin{tikzpicture}
  \node (img1)  {\includegraphics[scale=0.75]{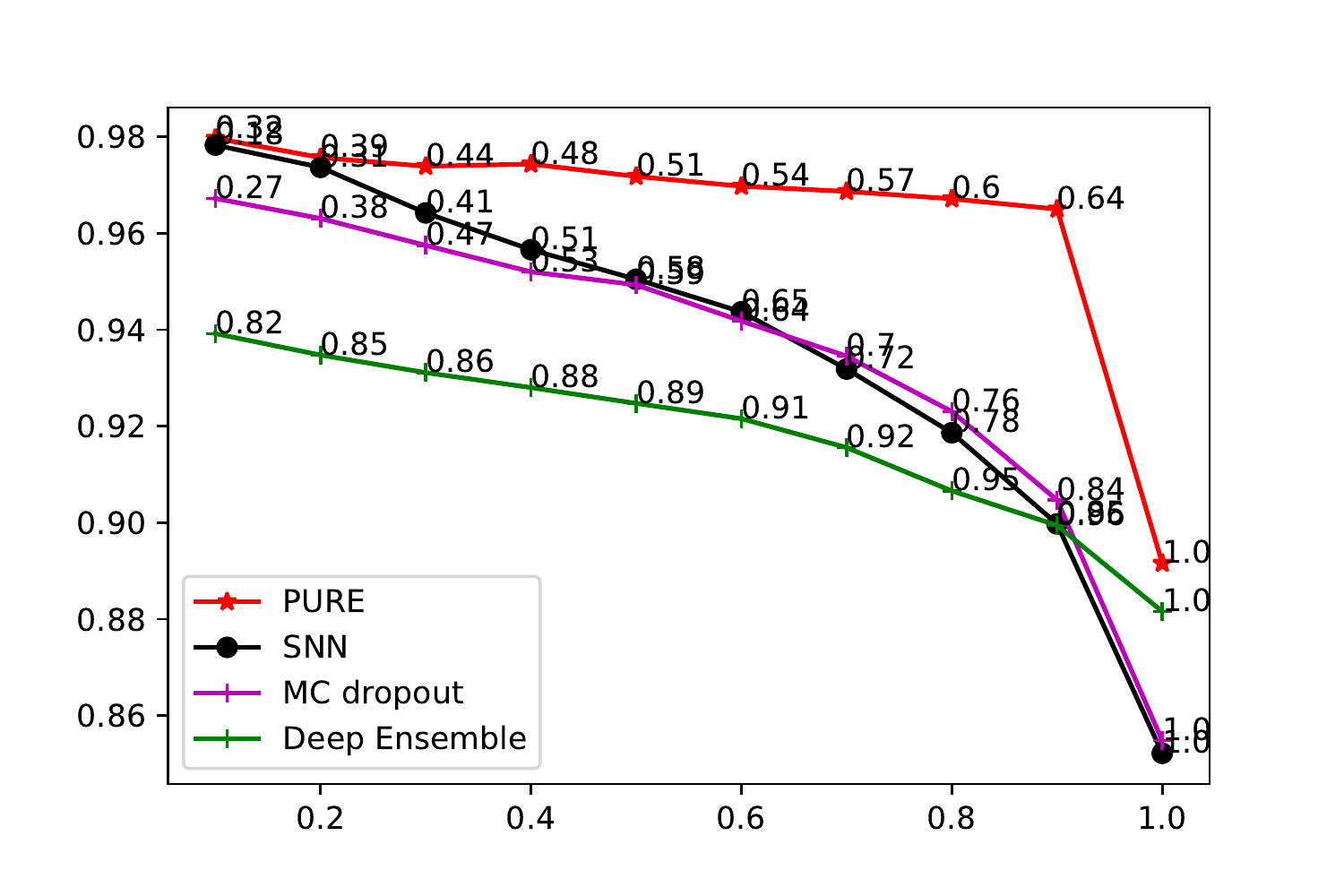}};
  \node[below=of img1, node distance=0cm, yshift=1.5cm] {Entropy Threshold};
  \node[left=of img1, node distance=0cm, rotate=90, anchor=center,yshift=-1.5cm] {Accuracy};
\end{tikzpicture}
	\caption{The change of accuracy for different models with respect to different entropy thresholds.}
	\label{fig:comparison_accuracy_all}
\end{figure}

Figure \ref{fig:comparison_accuracy_all} demonstrates the variation of the test accuracy for the \pure, SNN, MC dropout, and Deep Ensemble models as we change the thresholds for the normalized entropy for delegating predictions. 
The \pure~outperforms the SNN, MC dropout, and Deep Ensemble models at almost all points.
Its accuracy is higher than that of alternative models even when there are no delegated predictions and the disparity significantly increases as the predictions are filtered based on their entropy.
When we select entropy threshold as $0.4$, SNN achieves $95.6\%$ accuracy using $51\%$ of the data. For the same threshold, $53\%$ of the data used by MC dropout, and the accuracy value is $95.2\%$. Deep Ensemble obtains an accuracy of $93\%$ using $88\%$ of the data, whereas \pure~achieves $97.8\%$ of accuracy for $48\%$ of the predictions at the same threshold. 
Our observation is consistent with the literature, where the deep neural networks are criticized as being overconfident, hence misleading, when they make mistakes~\cite{hullermeier2019aleatoric}. 
One important aspect to note here is the distribution of the data over various entropy thresholds. In principle, we want to use the entropy threshold to decide if a decision will be delegated to the user. Consider \pure~in Figure \ref{fig:comparison_accuracy_all}. When entropy is $0.1$, \pure~will only classify $32\%$ of the data and delegate the remaining to the user. While this is a large percentage to delegate, it comes with the advantage of $98\%$ accuracy. If \pure~sets its entropy to $0.8$, then it will classify $60\%$ of the data and still yielding an accuracy of $98\%$. When it chooses to classify all the data, then the accuracy will drop to $89\%$. Contrast this ability to configure based on entropy to Deep Ensemble. With Deep Ensemble, even when the entropy is set to $0.1$, the agent will classify $82\%$ of the data itself, with low flexibility in delegating the choices to the user. Next, we study the accuracy changes of these models based on their data usage.

\begin{figure}[htb]
\centering
\begin{tikzpicture}
  \node (img2)  {\includegraphics[scale=0.75]{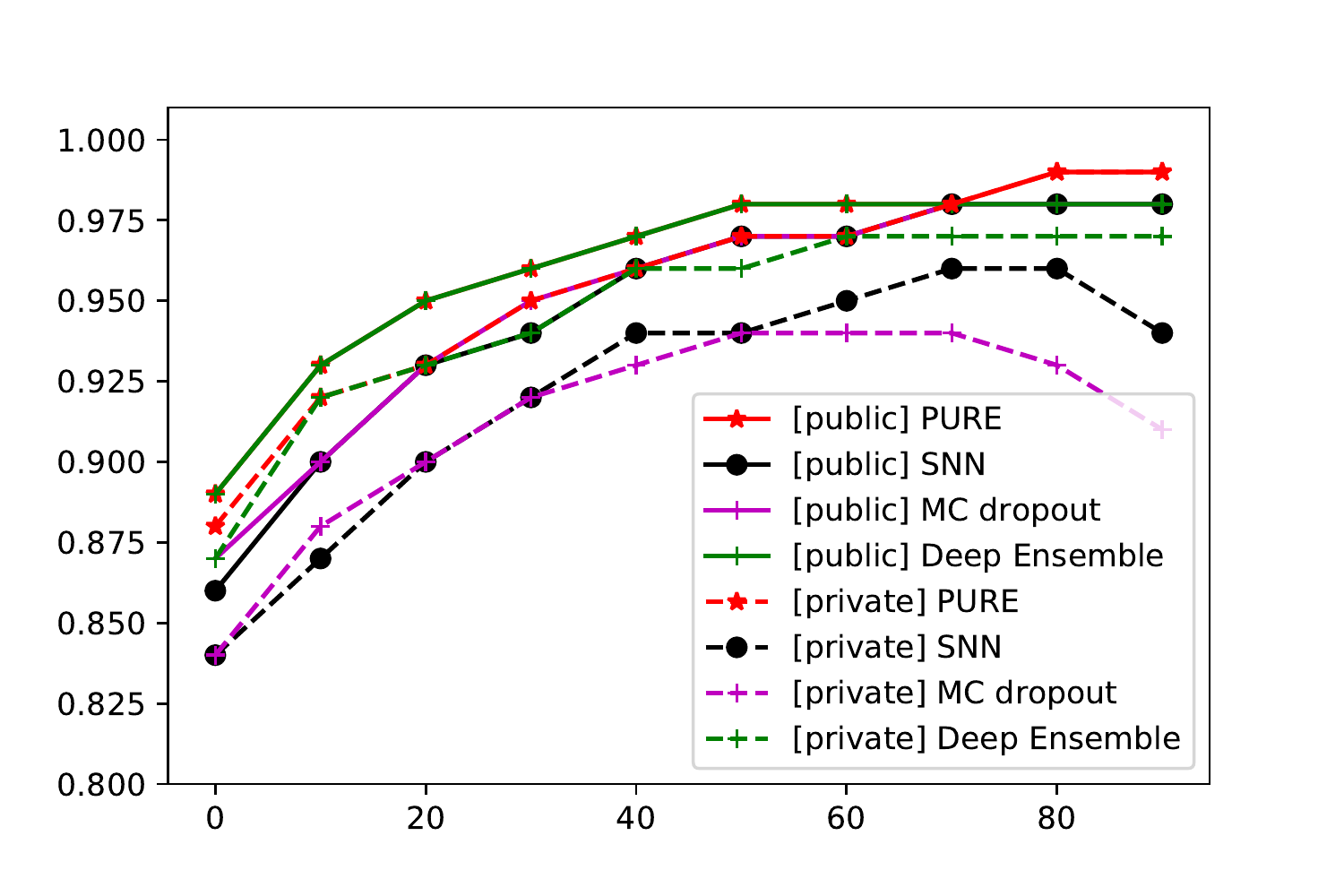}};
  \node[below=of img2, node distance=0cm, yshift=1.5cm] {Delegated Data Fraction [\%]};
  \node[left=of img2, node distance=0cm, rotate=90, anchor=center,yshift=-1.2cm] {F1-score};
\end{tikzpicture}
	\caption{F1-scores for the private and public classes relative to the percentage of delegated decisions.}
	\label{fig:comparison_f1_all}
\end{figure}

Figure \ref{fig:comparison_f1_all} plots how the F1-score changes for the \textit{public} and \textit{private} classes when \pure, SNN, MC dropout, and Deep Ensemble models delegate certain percentages of their most uncertain predictions based on the entropy.
The F1-scores of \pure~and Deep Ensemble models are better than SNN and MC dropout for each class and the gap is bigger for the private class. \pure~outperforms Deep Ensemble for both classes when the rate is higher than $0.7$. 
The F1-score of \pure~improves further and reaches $0.99$ for both private and public classes. However, the F1-score of the private class for SNN and MC dropout decreases when the most uncertain $80\%$ of the data is neglected and the remaining most certain $20\%$ is used for the calculation of the F1-score.
The decrease in the F1-score for the private class in Figure \ref{fig:comparison_f1_all} indicates that SNN and MC dropout is overconfident while \pure~can exploit its well-measured uncertainty to avoid wrong privacy decisions. 
With randomization tests \cite{noreen1989computer}, we can show that the improvements of \pure~over existing models is statistically significant (p-value $< 0.05$).
We answer RQ2 positively such that \pure~outperforms SNN, MC dropout, and Deep Ensemble by expressing uncertainty better.

\mypara{Personalized Misclassification Risk:}
Each user may have a significantly different cost for the misclassification of the private content.
In this section, we demonstrate the flexibility of \pure~for adapting users' perceived risk of such mistakes and its ability to avoid them by refraining from making privacy decisions when uncertain.

\begin{table*}[]
\centering
\begin{tabular}{|c|c|c|}
\hline  & \textbf{\begin{tabular}[c]{@{}c@{}}Non-Sensitive\\ ($R_{01} = R_{10} = 1$)\end{tabular}} & \textbf{\begin{tabular}[c]{@{}c@{}}Sensitive\\ ($R_{01} = 1, R_{10} = 10$)\end{tabular}} \\ \hline
\textbf{{[}Overall{]} Accuracy} & 0.89                                                                                   &   0.90                                                                                \\ \hline
\textbf{{[}Overall{]} Recall}   & 0.89                                                                                   &   0.90                                                                                 \\ \hline
\textbf{{[}Private{]} Recall}   & 0.86                                                                                    &   0.91                                                                                \\ \hline
\textbf{{[}Public{]} Recall}    & 0.92                                                                                   &  0.89                                                                              \\ \hline
\end{tabular}
\\[1ex]
\vspace{0.05in}
\caption{Results for two different risk personas.}
\label{tab:approaches_for_personas}
\end{table*}

For this purpose, we consider two broad categories of personas: \textit{non-sensitive} and \textit{sensitive}, which have a different risk matrix $R$. 
The non-sensitive user has the same perception of risk for the misclassification of private and public content, i.e., $R_{01}=R_{10}=1$.
This means that, for the non-sensitive user, the KL-term in the loss of the \pure~does not weigh the evidence for private and public categories differently.
On the other hand, for the sensitive user, misclassifying private content as the public is ten times more unacceptable, i.e., $R_{01}=1$ and $R_{10}=10$. This means that \pure~is significantly more penalized for making a wrong prediction in private class.

Table \ref{tab:approaches_for_personas} shows our results. For the non-sensitive persona, the results are as before. For the sensitive persona, the recall for the private category improves significantly at a cost of having lower recall for the public category. In other words, \pure~prefers to classify a content as private over public, when in doubt. This behaviour increases the number of predictions for the private category for the sensitive user (i.e., private recall increase from $0,86$ to $0.91$. While doing so, it does not sacrifice its overall recall and the accuracy. Notice that the $R$ value belongs to a user and can be adjusted as needed.

\mypara{Data Personalization:}
\pure~delegates a decision to its user when uncertain about a prediction. Ideally, we would like to minimize the number of times this happens while keeping the accuracy high. The personalization is meant to serve this purpose. In order to see if this is indeed achieved, we need to perform a comparative analysis where in the first round no personal data are used and in the second round the personal data are added. To realize this, we select three users who have annotated the most images as each annotator has annotated different number of images. 

\begin{itemize}
    \item \textbf{Round I:} Train a model without using personal data. 
    \item \textbf{Round II-Personal:} Tune a trained model in \textit{Round I} using personal data annotated by a user.    
    \item \textbf{Round II-Random:} Tune a trained model in \textit{Round I} using data annotated by others.
\end{itemize}

\begin{figure}
     \centering
     \begin{subfigure}{0.45\textwidth}
    \centering
    \begin{tikzpicture}
  \node (img1) {\includegraphics[width=\textwidth]{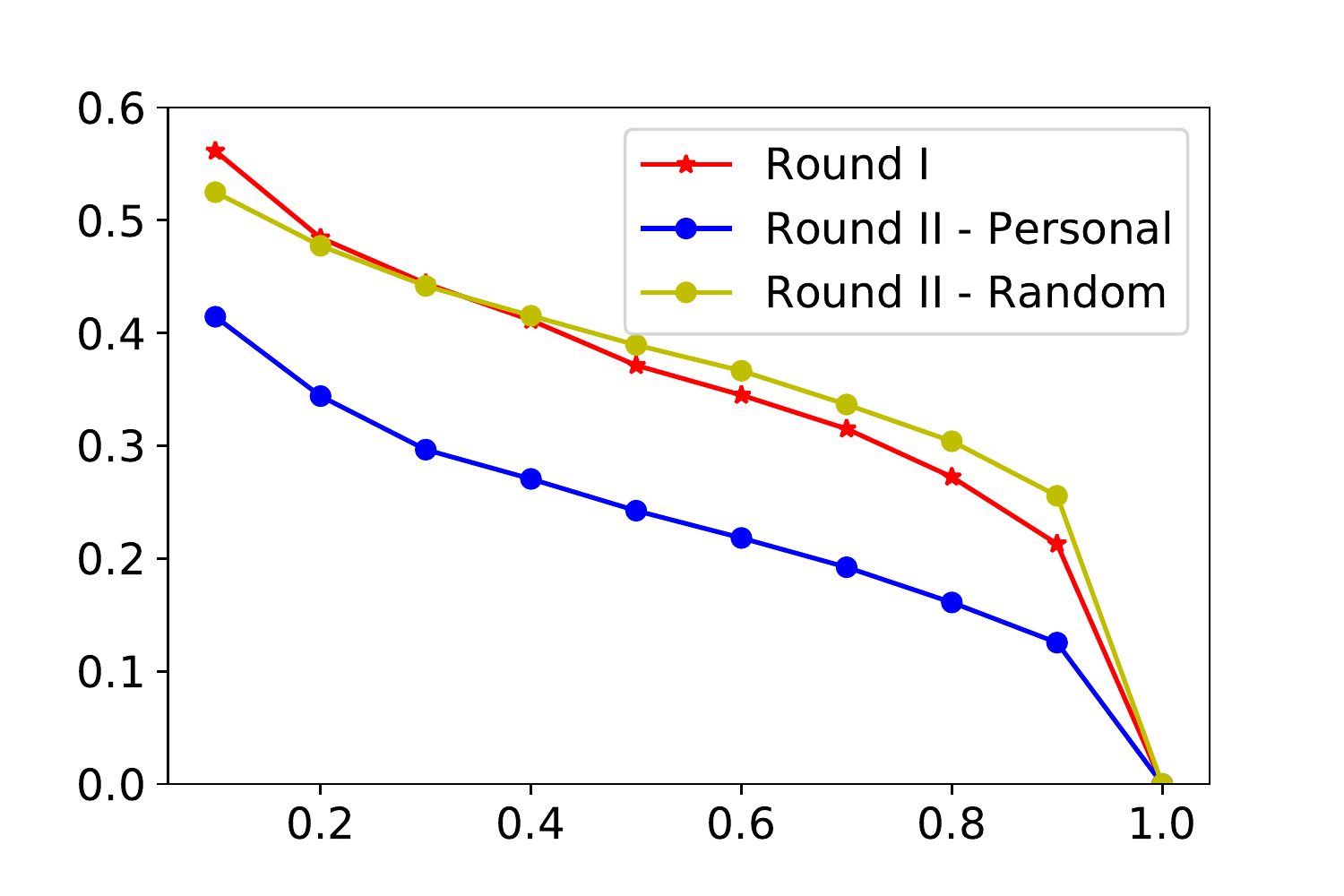}};
  \node[below=of img1, node distance=0cm, yshift=1.3cm] {Uncertainty Threshold};
  \node[left=of img1, node distance=0cm, rotate=90, anchor=center,yshift=-1.2cm] {Delegate [\%]};
 \end{tikzpicture}
\caption{User 1}
\label{fig:data_personalization_user1}
\end{subfigure}%
  \hspace{0.5in}
\begin{subfigure}{0.45\textwidth}
\centering
\begin{tikzpicture}
  \node (img2) {\includegraphics[width=\textwidth]{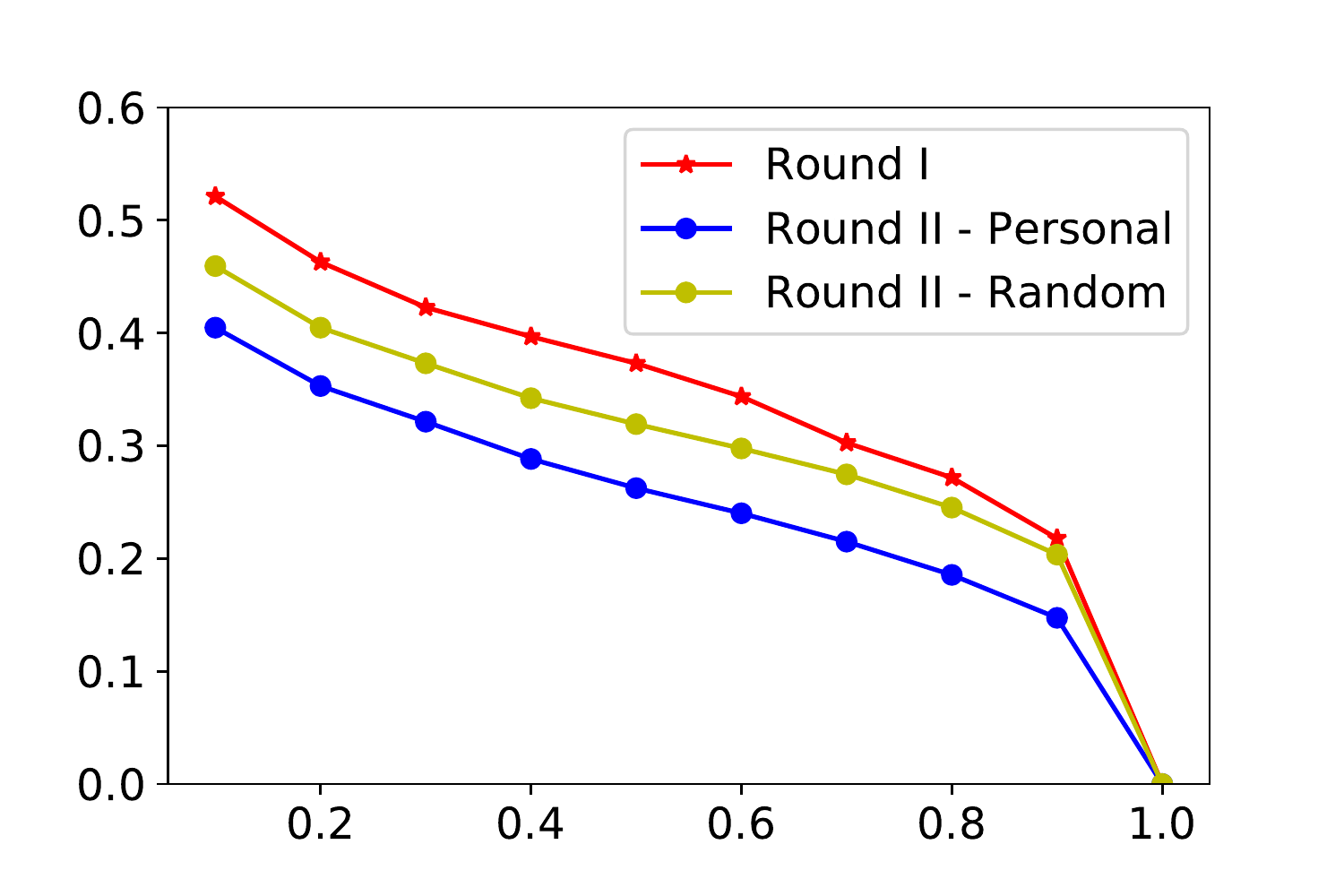}};
  \node[below=of img2, node distance=0cm, yshift=1.3cm] {Uncertainty Threshold};
  \node[left=of img2, node distance=0cm, rotate=90, anchor=center,yshift=-1.2cm] {Delegate [\%]};
\end{tikzpicture}
\caption{User 2}
\label{fig:data_personalization_user2}
\end{subfigure}%
  \hspace{0.5in}
\begin{subfigure}{0.45\textwidth}
\centering
\begin{tikzpicture}
 \node (img3) {\includegraphics[width=\textwidth]{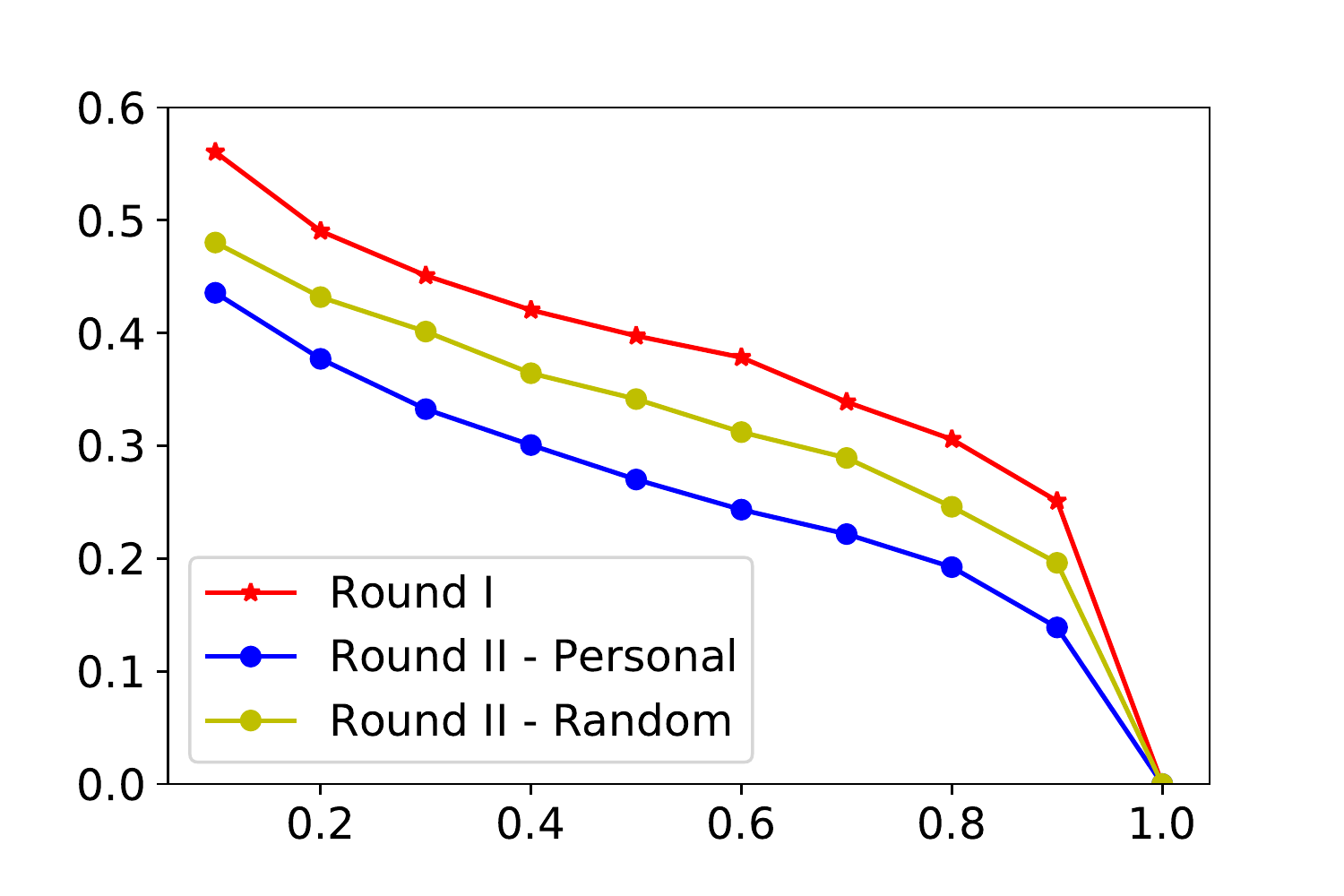}};
  \node[below=of img3, node distance=0cm, yshift=1.3cm] {Uncertainty Threshold};
  \node[left=of img3, node distance=0cm, rotate=90, anchor=center,yshift=-1.2cm] {Delegate [\%]};
 \end{tikzpicture}
\caption{User 3}
\label{fig:data_personalization_user3}
\end{subfigure}
\caption{The change of delegated samples for each round with respect to different uncertainty thresholds.}
\label{fig:data_personalization_users}
\end{figure}
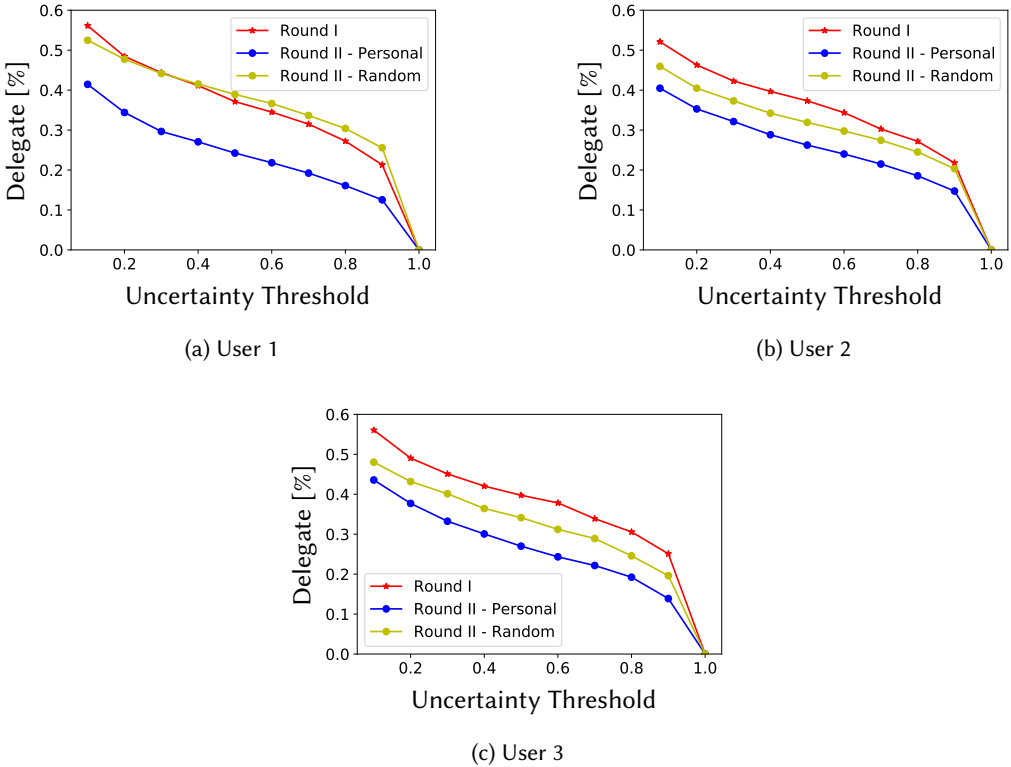

Figure \ref{fig:data_personalization_users} shows the change of samples (\%) that \pure~prefers to delegate decision of labels to the user.  We observe that \pure~delegates less to its user when it makes use of the personalization module.

A closer look in Table \ref{tab:data_personalization_assign_to_user} shows the ratio of samples at each round when uncertainty threshold is $0.7$. These samples belong to the top three users who annotate the most images. For instance, for the first user, \pure~delegates $32\%$ and $34\%$ of test samples after \textit{Round I} and \textit{Round II - Random}, respectively. However, only $19\%$ uncertain cases can be delegated to the user when we tune the trained model with personal data at Round II.  In light of these results, we answer RQ3 positively: \pure~can adjust its behavior based on the personal risk and expectations of its user as well as help the user deal with fewer decisions. 

\begin{table}[]
	\begin{tabular}{|c|c|c|c|}
		\hline
		$\theta=0.7$    & \textbf{Round I} & \textbf{Round II - Personal} & \textbf{Round II - Random} \\ \hline
		\textbf{User 1} & 0.32        & 0.19           & 0.34                    \\ \hline
		\textbf{User 2} & 0.30        & 0.21          & 0.27                 \\ \hline
		\textbf{User 3} & 0.34        & 0.22            & 0.29                     \\ \hline
	\end{tabular}
	\\[1ex]
	\vspace{0.05in}
	\caption{Ratios of prediction whose uncertainty values are greater than $0.7$ ($\theta$).}
	\label{tab:data_personalization_assign_to_user}
\end{table}

An interesting aspect to note is that the number of images use to personalize can affect the behavior of the data personalization approach. Given the limited number of annotators with predefined set of images, we currently cannot study such questions but it would be useful to provide bounds to guide users to personalize the assistant even further.

\section{Conclusion}
\label{sec:discussion}

This paper proposes a personal privacy assistant, \pure,~that helps its user make privacy decisions by recommending privacy labels (private or public) for given contents. \pure~is uncertainty-aware in that it captures the privacy ambiguity using uncertainty modeling and delegates decisions for ambiguous cases to its user. \pure~is personalized that it is capable of making a privacy decision by incorporating the user's risk of misclassification and using personally labeled data. Through its personalization, \pure~is also unobtrusive as it does not consult its user only when it is uncertain. Our experimental results show that \pure~obtains a high accuracy without even consulting its user at all. Our comparison with other models in the literature show that \pure~captures uncertainty well and that most of the content that it identifies as uncertain are the ones that it would have made an error if it were to classify them itself. Thus, the overall accuracy of \pure~steadily increases as it delegates uncertain cases to its user. Moreover, our results show that \pure~can indeed adjust its behavior based on the personal risk and expectations of its user and is able to decrease the delegations to its user by fine-tuning using personal data.

This paper opens up interesting directions for future research. Currently, we start with an uncertainty-aware model for privacy classification and enhanced it further with users' personalized risk for misclassification. An interesting direction for further research is to enable \pure~to have deeper interactions with the user. For example, it could interact with the user to obtain labels for the images that it is uncertain about and further enhance its ability for classification with this new personal data. Similarly, it could attempt to explain the uncertainty to its user so that the user can help in guiding \pure~in the right direction. Semantic information about the content, such as tags, could aid in the explanation. Another important direction is to enable interaction between different personal assistants to help create a collaborative environment for preserving privacy. Currently, we assume that the content that a personal assistant decides on belongs to its user alone. However, many times content, such as group images or co-edited documents, might belong to more than one user~\cite{ulusoy-yolum-22}. Extending \pure~to act collaboratively in such settings would be useful. Finally, a user's privacy preferences are many time relational. That is, a user might be fine with sharing a content with a friend but not with a colleague. Our current approach does not capture with whom the content is shared. It would be interesting to learn relation-based sharing behavior of users.

\section{Acknowledgments}
The first author is partially supported by the Scientific and Technological Research Council of Turkey (TÜBİTAK) and Turkish Directorate of Strategy and Budget under the TAM Project number 2007K12-873. This research was partially funded by the Hybrid Intelligence Center, a 10-year programme funded by the Dutch Ministry of Education, Culture and Science through the Netherlands Organisation for Scientific Research, \url{https://hybrid-intelligence-centre.nl}.



\end{document}